\documentclass[journal]{IEEEtran}

\ifCLASSINFOpdf

\else

\fi

\usepackage{graphicx}

\usepackage{fixltx2e}
\usepackage{url}
\usepackage{xspace}
\usepackage{booktabs}
\usepackage{balance}
\usepackage{paralist}
\usepackage{multirow}
\usepackage{amsmath}
\usepackage[table]{xcolor}
\usepackage[linesnumbered,ruled]{algorithm2e}

\usepackage[caption=false,font=footnotesize]{subfig}
\definecolor{Gray}{gray}{1} 

\graphicspath{{pics/}}

\begin{document}

\title{Learning One Class Representations for Face Presentation Attack Detection using Multi-channel Convolutional Neural Networks}
%
%


\author{Anjith~George, \IEEEmembership{Member,~IEEE}~and~S\'ebastien~Marcel,~\IEEEmembership{Senior~Member,~IEEE}
\thanks{A. George and S. Marcel are in Idiap Research Institute, Centre du Parc, Rue Marconi 19, CH - 1920, Martigny, Switzerland.
(e-mail: \{anjith.george,sebastien.marcel\}@idiap.ch)}}


\maketitle

\begin{abstract}
Face recognition has evolved as a widely used biometric modality. However, its vulnerability against presentation attacks poses a significant security threat. Though presentation attack detection (PAD) methods try to address this issue, they often fail in generalizing to unseen attacks. In this work, we propose a new framework for PAD using a one-class classifier, where the representation used is learned with a Multi-Channel Convolutional Neural Network (\textit{MCCNN}). A novel loss function is introduced, which forces the network to learn a compact embedding for \textit{bonafide} class while being far from the representation of attacks. A one-class Gaussian Mixture Model is used on top of these embeddings for the PAD task.  The proposed framework introduces a novel approach to learn a robust PAD system from \textit{bonafide} and available (known) attack classes. This is particularly important as collecting \textit{bonafide} data and simpler attacks are much easier than collecting a wide variety of expensive attacks. The proposed system is evaluated on the publicly available \textit{WMCA} multi-channel face PAD database, which contains a wide variety of 2D and 3D attacks. Further, we have performed experiments with \textit{MLFP} and \textit{SiW-M} datasets using RGB channels only. Superior performance in unseen attack protocols shows the effectiveness of the proposed approach. Software, data, and protocols to reproduce the results are made available publicly.

\end{abstract}

\begin{IEEEkeywords}
Presentation Attack Detection, Convolutional Neural Network, Face Recognition, Anti-spoofing, Reproducible Research, Unseen Attack Detection.
\end{IEEEkeywords}

\IEEEpeerreviewmaketitle

\section{Introduction}

\IEEEPARstart{F}{ace} recognition has proved to be a beneficial modality for biometric authentication. One of the main reasons for the widespread use of face recognition systems is its non-intrusive nature of acquisition and ease of use \cite{jain2011handbook}. Face recognition systems have matured a lot in recent years, and several approaches have reported human parity in the identification rate in `in the wild' conditions \cite{learned2016labeled}. However, a critical security issue undermining the widespread use of face recognition technology is its vulnerability to presentation attacks (a.k.a spoofing attacks) \cite{handbook2}, \cite{ISO}.

\begin{figure}[t!]
\centering
\includegraphics[width=0.99\linewidth]{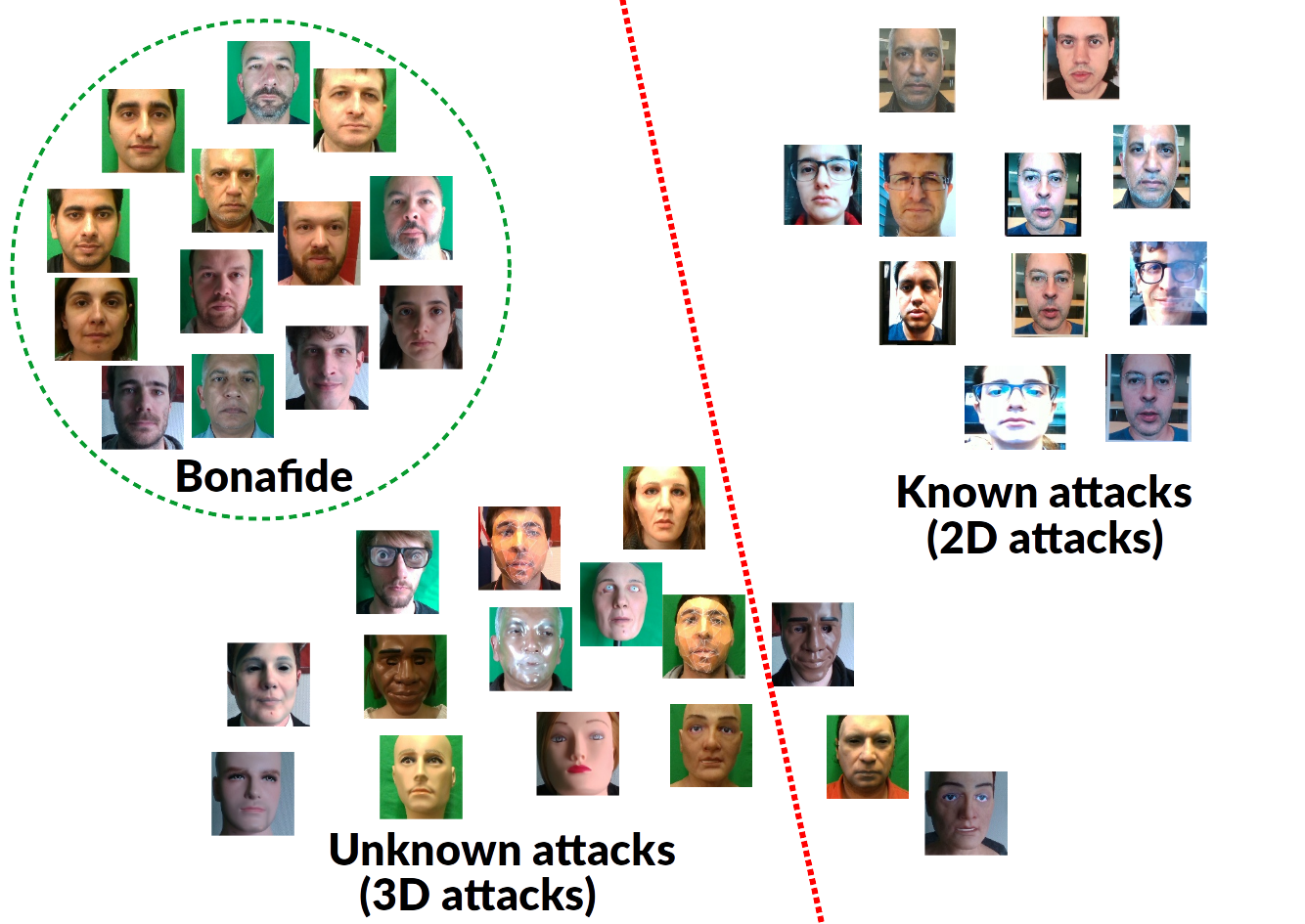}
\caption{Illustration of the embedding space with known and unknown attack classes. The red dotted line shows the decision boundary learned when only \textit{bonafide} and known attacks are present in the training set, this results in misclassification of unknown attacks as \textit{bonafide}. If a one class decision boundary (green-dotted lines) is learned, then both known and unknown attacks can be classified correctly. }
\label{fig:embedding_space}
\end{figure}

Presentation attack refers to an attack using an instrument with the intention to affect the normal operation of the biometric system. Often, features such as color, texture \cite{boulkenafet2015face}, \cite{maatta2011face}, motion \cite{anjos2011counter}, and physiological cues \cite{ramachandra2017presentation}, \cite{heusch2019remote} and CNN based methods \cite{george2019deep} are used for detection of attacks like 2D prints and replays. However, detection of sophisticated attacks like 3D masks and partial attacks are challenging and poses a serious threat to the reliability of face recognition systems. Most of the presentation attack detection (PAD) methods available in prevailing literature try to solve the problem for a limited number of presentation attack instruments and on visible spectrum images \cite{handbook2}. Though some success has been achieved in addressing 2D presentation attacks, performance of the algorithms in realistic 3D masks and other kinds of attacks is poor. With the increase in quality of attack instruments, it becomes harder to discriminate between \textit{bonafide} and PAs in the visible spectrum alone.  Moreover, considering a real-world situation with a wide variety of 2D, 3D, and partial attacks, PAD in visual spectra alone is challenging and inadequate for security-critical applications. Partial attacks refer to attacks where the attack instrument covers only a part of the face. These attacks are much harder to detect as they appear similar to \textit{bonafide} in most of the face regions, and they can fool holistic liveliness detection systems easily. Multi-channel methods have been proposed as an alternative \cite{raghavendra2017extended}, \cite{steiner2016reliable}, \cite{bhattacharjee2017you}, \cite{george_mccnn_tifs2019,george2020face,heusch2020deep,george2020face,nikisins2019domain}, since they use complementary information from different channels to improve the discrimination between \textit{bonafide} and attacks. In the multi-channel scenario, the additional channels used can be any modality which can provide complementary representation such as depth, infrared, and thermal channels. Multi-channel PAD approaches are more promising in the context of a wide variety of attacks since they make PAD systems harder to fool.

Even with the use of multiple channels, one of the main issues with PAD is its poor generalization to unseen attacks \cite{george_mccnn_tifs2019}. This is particularly important, since at the time of developing a PAD system, anticipating all possible attacks is impossible. Malicious attackers can always come up with new attacks to fool the PAD systems. In such situations, PAD systems which are robust against unseen attacks are of paramount importance. Moreover, while it is comparatively easy to collect data for attacks like 2D prints and replays, making replicas of challenging presentation attack instruments (PAI) like silicone mask are often very costly \cite{Bhattacharjee:256262} and resource-intensive. In this context, it will be ideal to have a framework which can be trained with \textit{bonafide} alone, or with a combination of \textit{bonafide} and easy to manufacture PAIs.

In real-world scenarios, it can be assumed that all presentation attacks are unseen, as it is not possible to foretell all the variations a PAD system could encounter a priori. A toy example of the decision boundary in an unseen attack scenario is illustrated in Fig. \ref{fig:embedding_space}. Performances in typical PAD databases may not be representative of the performance of a PAD system in real-world conditions. This necessitates the PAD algorithms to be robust against unseen attacks. Since it is easy  (in effort and cost) to collect data from more straightforward attacks compared to complex PAIs, we try to learn the representation leveraging the information from PA classes which are available at the training stage (while not over-fitting on the available attacks). To achieves this, we propose a one-class classifier based framework, where the feature representation is learned with a CNN to have discriminative properties. The core of the framework is a multi-channel CNN trained to learn the embedding using a specific loss function. The proposed approach aims at learning a compact representation for the \textit{bonafide} class while leveraging the discriminative information for PAD task.

The main contributions of the paper are listed below.

\begin{itemize}
  \item A novel multi-channel one-class classifier-based approach is proposed for unseen attack detection.
  \item A novel loss function is proposed which learns a compact and discriminative representation of the face for PAD task, leveraging the information provided from known attacks.

\end{itemize}

The features used in the one class classifier are learned with a multi-channel CNN framework. The proposed approach was evaluated in \textit{known} and \textit{unseen} attack protocols in \textit{WMCA} database containing a wide variety of 2D and 3D attacks, and performed significantly better than baselines in \textit{unseen} protocols. We have also performed experiments using RGB channel in \textit{MLFP} and \textit{SiW-M} datasets.

Additionally, the source code and protocols to reproduce the results are made available publicly and are accessible at the following link \footnote{Source code: \url{https://gitlab.idiap.ch/bob/bob.paper.oneclass_mccnn_2019}}.

The rest of the paper is organized as follows. Section 2 describes the related work with a particular focus on unseen attack detection. Section 3 outlines the proposed framework. Extensive evaluations, comparison with baseline methods, and ablation studies are shown in section 4. Section 5 discusses the importance of the results, and Section 6 presents the conclusions.

\section{Related work}

Majority of the literature in face PAD is mainly focused on 2D attacks and uses feature-based methods \cite{boulkenafet2015face}, \cite{maatta2011face},  \cite{anjos2011counter},\cite{ramachandra2017presentation}, \cite{heusch2019remote} or CNN based methods. Recently, CNN based methods have been more successful as compared to feature-based methods \cite{liu2018learning}, \cite{george2019deep}, \cite{atoum2017face}, \cite{shao2017deep}. These methods usually leverage the quality degradation during `recapture' and are often useful only for the detection of attacks like 2D prints and replays. Sophisticated attacks like 3D masks are more challenging and pose serious threat to the reliability of face recognition systems.

Most of these methods handle the PAD problem as binary classification, which results in classifiers over-fitting to the known attacks resulting in poor generalization to unseen attacks. We focus the further discussion on the detection of unseen attacks. However, it is imperative that methods working for unseen attacks must perform accurately for known attacks as well. One naive solution for such a task is one-class classifiers (OCC). OCC provides a straightforward way of handling the unseen attack scenario by modeling the distribution of the \textit{bonafide} class alone.

Arashloo \textit{et al}.\cite{arashloo2017anomaly} and Nikisins \textit{et al}. \cite{nikisins2018effectiveness} have shown the effectiveness of one class methods against unseen attacks. Even though these methods performed better than binary classifiers in an unseen attack scenario, the performance in known attack protocols was inferior to that of binary classifiers. Xiong \textit{et al}. \cite{xiong2018unknown} proposed unseen PAD methods using auto-encoders and one class classifiers with texture features extracted from images.  However, the performance of the methods compared to recent CNN based methods is very poor. CNN based methods outperform most of the feature-based baselines for PAD task. Hence there is a clear need of one class classifiers or anomaly detectors in the CNN framework. One of the drawbacks of one class model is that they do not use the information provided by the known attacks. An anomaly detector framework which utilizes the information from the known attacks could be more efficient.

Perera and Patel \cite{perera2019learning} presented an approach for one-class transfer learning in which labelled data from an unrelated task is used for feature learning. They used two loss functions, namely
descriptive loss, and compactness loss to learn the representations. The data from the class of interest is used to calculate the compactness loss whereas an external multi-class dataset is used to compute the descriptive loss. Accuracy of the learned model in classification using another database is used as the descriptive loss. However, in the face PAD
problem, this approach would be challenging since the \textit{bonafide} and attack classes appear very similar.

Fatemifar \textit{et al}. \cite{fatemifar2019combining} proposed an approach to ensemble multiple one-class classifiers for improving the generalization of PAD. They introduced a class-specific normalization scheme for the one class scores before fusion. Seven regions, three one class classifiers and representations from three CNNs were used in the pool of classifiers. Though their method achieved better performance as compared to client independent thresholds, the performance is inferior to CNN based state of the art methods. Specifically, many CNN based approaches have achieved 0\% HTER in Replay-Attack and Replay-Mobile datasets. Moreover, the challenging unseen attack scenario is not evaluated in this work.

P{\'e}rez-Cabo \textit{et al}. \cite{perez2019deep} proposed a PAD formulation from an anomaly detection perspective.
A deep metric learning model is proposed, where a triplet focal loss is used as a regularization for `metric-softmax', which forces the network to learn discriminative features. The features learned in such a way is used together with an SVM with RBF kernel for classification. They have performed several experiments on an aggregated RGB only datasets showing the improvement made by their proposed approach. However, the analysis is mostly limited to RGB only models and 2D attacks. Challenging 3D and partial attacks are not considered in this work. Specifically, the effectiveness in challenging unknown attacks (2D vs 3D) is not evaluated.

Recently, Liu \textit{et al}. \cite{Liu_2019_CVPR} proposed an approach for
the detection of unknown spoof attacks as Zero-Shot Face Anti-spoofing (ZSFA).  They proposed a Deep Tree Network (DTN) which partitions the attack samples into semantic sub-groups in an unsupervised manner. Each tree node in their network consists of a Convolutional Residual Unit (CRU) and a Tree Routing Unit (TRU).  The objective is to route the unknown attacks to the most proper leaf node for correctly classifying it. They have considered a wide variety of attacks in their approach and their approach achieved superior performance compared to the considered baselines.

Jaiswal \textit{et al}. \cite{jaiswal2019ropad} proposed an end to end deep learning model for PAD which used unsupervised adversarial invariance. In their method, the discriminative information and nuisance factors are disentangled in an adversarial setting. They showed that by retaining only discriminative information, the PAD performance improved for the same base architecture. Mehta \textit{et al}. \cite{mehtacrafting} trained an Alexnet model with a combination of cross-entropy and focal losses. They extracted the features from Alexnet and trained a two-class SVM for PAD task. However, results in challenging datasets such as OULU and SiW were not reported.

Recently Joshua and Jain \cite{engelsma2019generalizing} utilized multiple GANs for spoof detection in fingerprints. Their method essentially consisted of training a DCGAN \cite{radford2015unsupervised} using only the \textit{bonafide} samples. At the end of the training, the generator is discarded, and the discriminator is used as the PAD classifier. They combined the results from different GANs operating on different features. However, this approach may not work well for face images as the recaptured images look very similar to the \textit{bonafide} samples.

In safety critical applications, extended range methods have been proposed over the years \cite{raghavendra2017extended}, \cite{erdogmus2014spoofing}, \cite{steiner2016reliable}, \cite{dhamecha2013disguise}, \cite{agarwal2017face}, \cite{Bhattacharjee:256262}, \cite{bhattacharjee2017you}, \cite{george_mccnn_tifs2019} to achieve reliable PAD performance. Even these methods fail in generalizing to unseen attacks.

Wang \textit{et al}. \cite{wang2019multi} proposed multimodal face presentation attack detection with a ResNet based network using both spatial and channel attentions.
Specifically, the approach was tailored for the \textit{CASIA-SURF} \cite{zhang2018casia} database which contained RGB, near-infrared and depth channels. The proposed model is a multi-branch model where the individual channels and fused data are used as inputs. Each input channel has its own feature extraction module and the features extracted are concatenated in a late fusion strategy. Followed by more layers to learn a discriminative representation for PAD. The network training is supervised by both center loss and softmax loss. One key point is the use of spatial and channel attention to fully utilize complementary information from different channels. Though the proposed approach achieved good results in the \textit{CASIA-SURF} database, the challenging problem of unseen attack detection is not addressed.

Parkin \textit{et al}. \cite{parkin2019recognizing} proposed a multi-channel face PAD network based on ResNet. Essentially, their method consists of different ResNet blocks for each channel followed by fusion. Squeeze and excitation modules (SE) are used before fusing the channels, followed by remaining residual blocks. Further, they add aggregation blocks at multiple levels to leverage inter-channel correlations. Their approach achieved state of the art results in \textit{CASIA-SURF} \cite{zhang2018casia} database. However, the final model presented in is a combination of 24 neural networks trained with different attack specific folds, pre-trained models and random seeds, which would increase the computation greatly.

From the discussions above, it can be seen that one class classifiers could be a good alternative for binary classification in PAD task. However, the features used for one class classifiers should be discriminative and compact to outperform binary classification.

\section{Proposed method}

From a practical viewpoint, it is not possible to anticipate all the possible types of attacks and to have them in the training set. This, in turn, make the PAD task an unseen classification problem in a broad sense. In general, we can even consider attacks coming from different replay devices as unseen attacks. Typically, one class classifiers are well suited for such outlier detection tasks. However, in practice, the performance of one class classifiers are inferior compared to binary classifiers for known attacks, since they do not leverage useful information from the known attacks. Ideally, the PAD system should perform well in both known and unseen attack scenarios.

Clearly, there is a necessity of a method which can learn a compact one class representation while utilizing the discriminative information from known attacks. While the collection of attacks could be difficult and costly, collecting \textit{bonafide} samples are rather easy. A new classification strategy is required to handle the realistic scenario where a limited variety of attack classes are available.

Though one class classifiers (\textit{OCC}) offers a way to model the \textit{bonafide} class, the efficient use of \textit{OCC} requires the feature representation to be compact while containing discriminative information for PAD task.
In the proposed framework, we use a CNN based approach to learn the feature representation. A novel loss function is proposed to learn a representation of \textit{bonafide} samples leveraging the known attack classes.

\subsection{Formulation of One Class Contrastive Loss (OCCL)}

Consider a typical CNN architecture for PAD, where the output layer contains one node and the loss function used is  Binary Cross Entropy (\textit{BCE}), which is defined as:

\begin{equation}
\mathcal{L}_{BCE}=-{(y\log(p) + (1 - y)\log(1 - p))}
\end{equation}
where $y$ is the ground truth, ($y=0$ for attack and $y=1$ for \textit{bonafide}) and $p$ is the probability.

When trained only with \textit{BCE} loss, the network learns a decision boundary based on the \textit{bonafide} and attacks present in the training set. However, it may not generalize when encountered with an unseen attack in the test time as it could be over-fitted to attacks which are `known' from the training set.

To overcome this issue, we propose the `One-Class Contrastive Loss' (\textit{OCCL}) function which operates on the embedding layer. Proposed One-Class Contrastive Loss (\textit{OCCL}) function is used as an auxiliary loss function in conjunction with binary cross-entropy loss. The feature map obtained from the penultimate layer of the CNN is used as the embedding. The loss function is inspired from  center-loss \cite{wen2016discriminative} and contrastive loss \cite{hadsell2006dimensionality}, which are usually used in the face recognition applications.

\begin{figure}[ht!]
     \centering
         \includegraphics[width=1.0\linewidth]{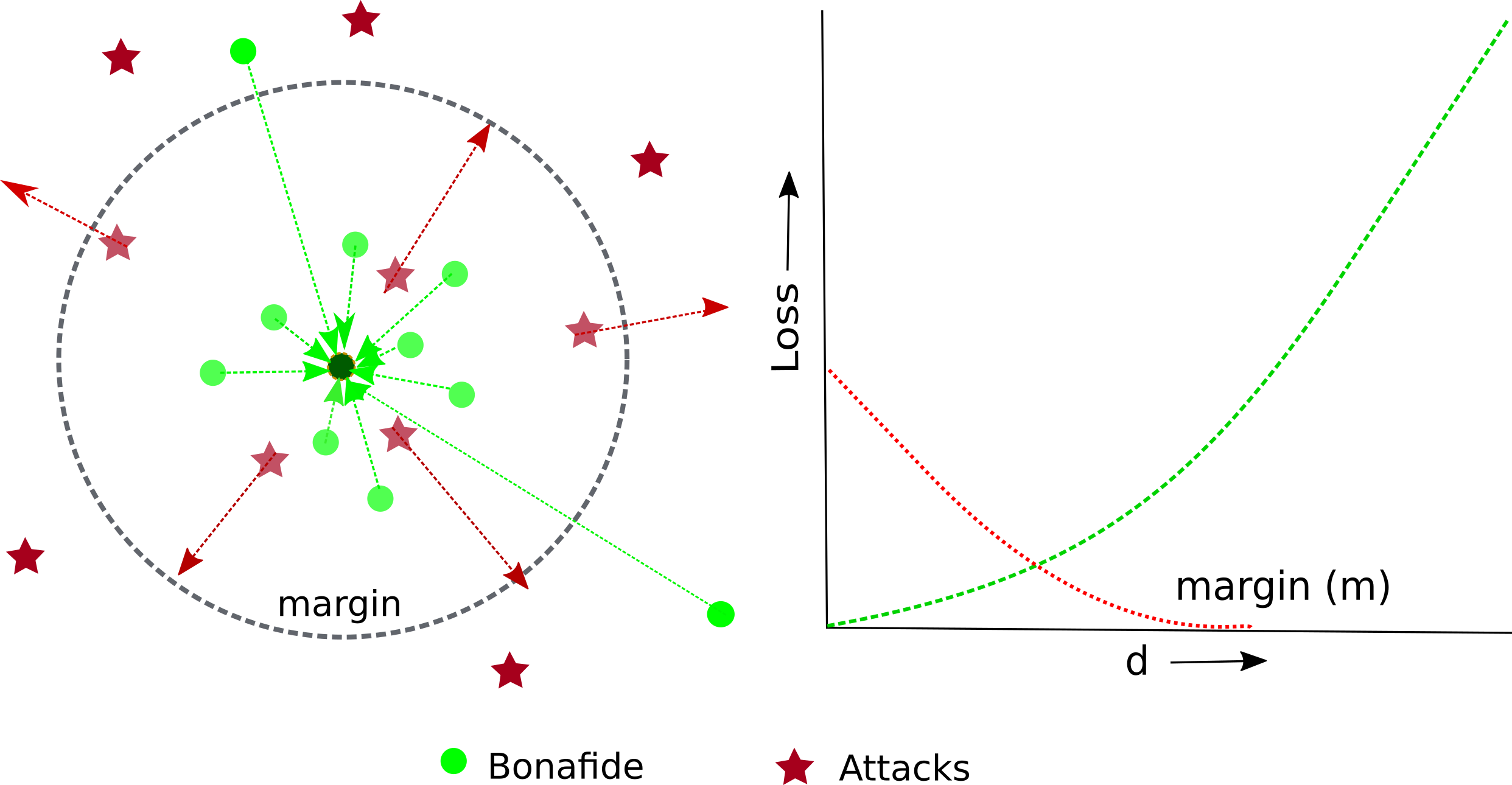}
\caption{Loss functions acting on the embedding space, left) \textit{bonafide} representations are pulled closer to the center of \textit{bonafide} class (green), while the attack embeddings(red) are forced to be beyond the margin. The attack samples outside the margin does not contribute to the loss, right) The loss as a function of distance from the \textit{bonafide} center.}
\label{fig:embedding_force}
\end{figure}

In face recognition applications, center loss is used as an additional auxiliary loss function, the task of the center loss is to minimize the distance of the embeddings from their corresponding class centers. The center loss is defined as:
\begin{equation}
\mathcal{L}_{center}=\frac{1}{2}\sum_{i=1}^m \|x_i-c_{y_i}\|_2^2
\end{equation}

Where $L_{center}$ denotes the center loss, $m$ the number of training samples in a mini-batch, $x_i \in R_d$ denotes the $i^{th}$ training sample, $y_i$ denotes the label, and $c_{y_i} $ denotes the $y^{th}_i$ class center in the embedding space.

The main issue with center loss in the PAD application is that the loss function penalizes for large intra-class distances and does not care about the inter-class distances. Contrastive center loss \cite{qi2017contrastive} tries to solve this issue by adding the distance between classes (inter-class) in the formulation. However, for the PAD problem, modeling the attack class as a cluster and finding a center for the attack class is not trivial. The attacks could be of different categories: 2D, 3D, and partial attacks, and it is not ideal forcing them to cluster together in the embedding space. It is only necessary to have the embeddings of attacks far from \textit{bonafide} cluster in the embedding space. Hence, we put the compactness constraint only on the \textit{bonafide} class, while forcing the embeddings of PAs to be far from that of \textit{bonafide}.

\begin{figure*}[h!]
     \centering
         \includegraphics[width=0.7\linewidth]{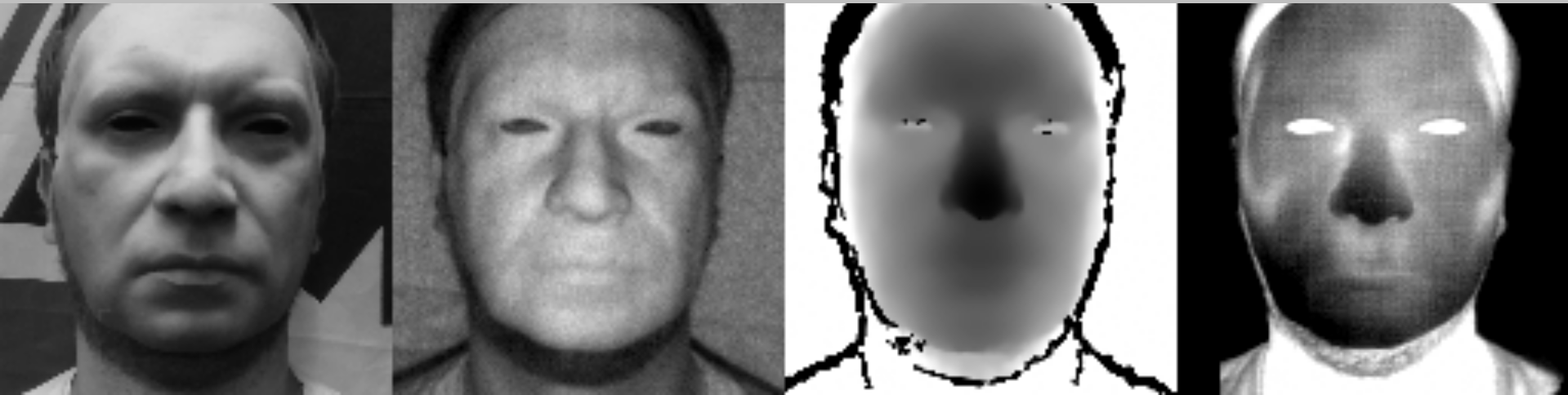}

\caption{Preprocessed images from a rigid mask attack; channels showed are gray-scale, infrared, depth, and thermal, respectively. Channels were preprocessed with face detection, alignment and normalization.}
\label{fig:preprocessed}
\end{figure*}

To formulate the loss function, we start with the equation for contrastive loss function proposed by Lecun \textit{et al}. \cite{hadsell2006dimensionality}.
\begin{equation}
\begin{split}
\mathcal{L}_{Contrastive}(W, Y, X^{1}, X^{2})= & (1-Y)\frac{1}{2}D_W^{2}  \\
            & + Y\frac{1}{2}{max(0, m-D_W)}^{2}
\end{split}
\end{equation}

Where $W$ is the network weights, $X^{1}, X^{2}$ are the pairs and $Y$ the label of the pair, i.e., whether they belong to the same class or not. $m$ is the margin, and $D_W$ is the distance function between two samples. The data is provided as pairs $(X^{1}, X^{2})$ and the distance function $D_W$ can be computed as the Euclidean distance.

\begin{equation}
D_W=\sqrt{\|X^{1}-X^{2}\|_2^2}
\end{equation}

Now, in our loss formulation, the critical difference is how we define $D_W$. In the original contrastive loss, $D_W$ is the distance between samples. In our case, we need the representation of \textit{bonafide} samples to be compact in an embedding space. At the same time, we want to maximize the distance between \textit{bonafide} cluster and attack samples in the embedding space. This can be achieved by defining $DC_W$ to be the distance from the center of \textit{bonafide} class as follows.

\begin{equation}
DC_W=\sqrt{\|X^{i}-c_{BF}\|_2^2}
\end{equation}

Where $X^{i}$ is the embedding for $i^{th}$ sample, and $c_{BF}$ is the center of \textit{bonafide} class in the embedding space.

The center of the \textit{bonafide} class is updated in every mini-batch during training as follows.
\begin{equation}
c_{BF}= \hat{c}_{BF} (1-\alpha) + \alpha \frac{1}{N}\sum_{i=1}^{N}e_{i}
\end{equation}

Where $c_{BF}$ and $\hat{c}_{BF}$ denotes the new and old \textit{bonafide}-centers. $\alpha$ is a scalar which prevents sudden changes in the class centers in mini-batch.  $e_{i}$ denotes the difference between embeddings for the \textit{bonafide} samples in the current mini-batch compared to the previous center, and $N$ denotes the number of \textit{bonafide} samples in the mini-batch.

Combining the equations, our auxiliary loss function becomes:

\begin{equation}
\begin{split}
\mathcal{L}_{OCCL}(W, Y, X)=  & Y\frac{1}{2}DC_W^{2} \\
            & + (1-Y)\frac{1}{2}{max(0, m-DC_W)}^{2}
\end{split}
\end{equation}
Where $DC_W$ denotes the Euclidean distance between the samples and the \textit{bonafide} class center, $Y$ denotes the ground truth, i.e., $Y=0$ for attacks and $Y=1$ for \textit{bonafide} (note the change in labels from the standard notation due to the ground truth convention). It is to be noted that, the proposed loss function \textit{\textbf{does not}} require pairs of samples, which is a requirement in usage of contrastive loss. This makes it easier to train the model without requiring an explicit selection of pairs during training.

This auxiliary loss makes the representation of \textit{bonafide} compact pushing it closer to the center of \textit{bonafide} class and penalizes attack samples which are closer than the margin $m$. Attack samples which are farther than the margin $m$ are not penalized. An illustration of the loss functions acting on the embeddings of \textit{bonafide} and attack samples are shown in Fig. \ref{fig:embedding_force}.

We combine the proposed loss function with standard binary cross entropy for training. The combined loss function to minimize is given as:

\begin{equation}
\mathcal{L} = (1-\lambda)\mathcal{L}_{BCE} + \lambda \mathcal{L}_{OCCL}
\end{equation}
 Where $\mathcal{L}$ denotes the total loss for the CNN. $\mathcal{L}_{BCE}$ and $\mathcal{L}_{OCCL}$ denotes the binary cross entropy, and one-class contrastive loss  respectively.  $\lambda$ denotes a scalar value to set the weight for each loss functions. In our experiments we set the value of $\lambda$ as $0.5$.

The combined loss function $\mathcal{L}$ tries to learn a decision boundary between the available attacks and \textit{bonafide} while the auxiliary loss tries to make the feature representation of the \textit{bonafide} compact in the embedding space. We expect the decision boundary learned in this fashion to be more robust in unseen attacks compared to the network learned only with \textit{BCE}. The embedding obtained in this manner is used with a one-class classifier for the PAD task.
\subsection{ Components of the proposed framework}

Different stages of the proposed framework are described below.
\subsubsection{Preprocessing}

\begin{figure*}[t!]
     \centering
         \includegraphics[width=1.0\textwidth]{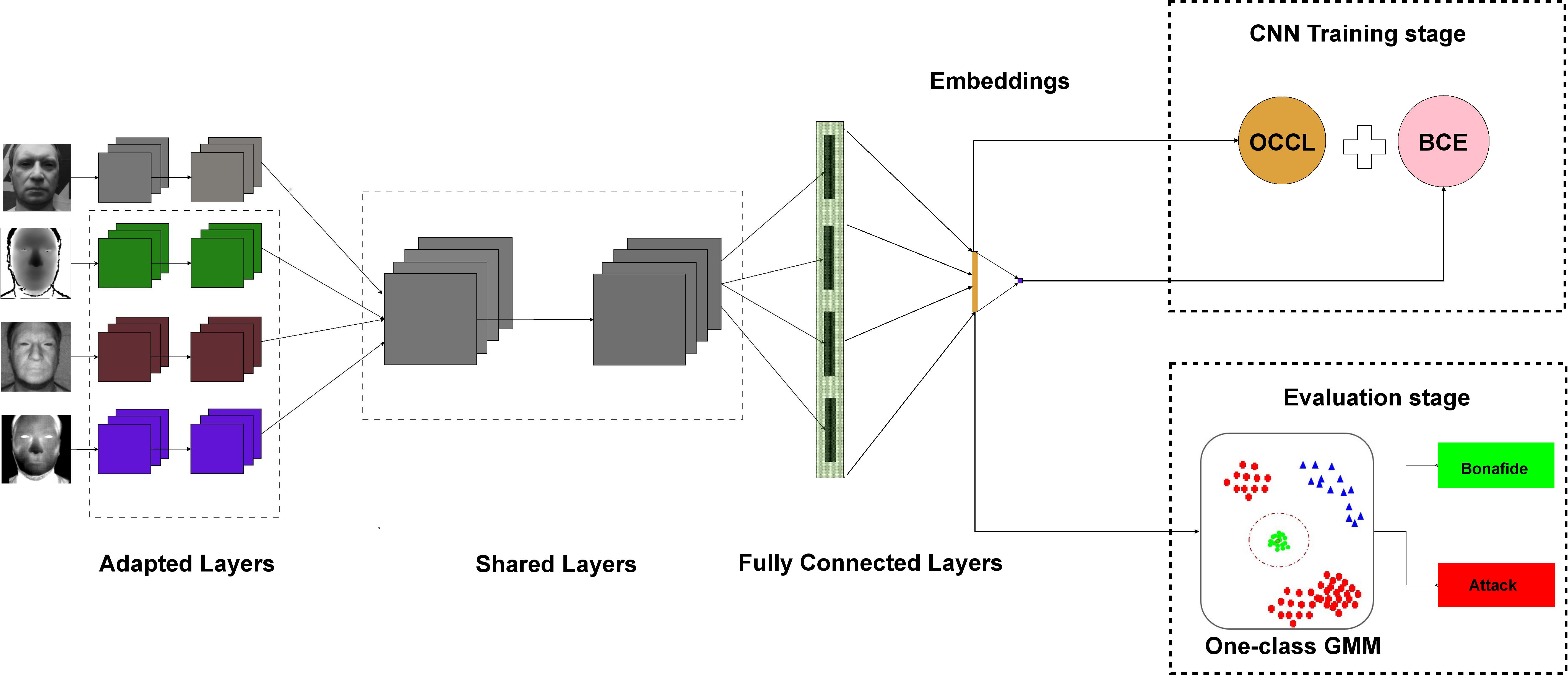}

\caption{Schematic diagram of the proposed framework. The CNN architecture is trained with two losses and then used as a fixed feature extractor with frozen weights. The one-class GMM is trained using the embeddings obtained from \textit{bonafide} class alone. }
\label{fig:arch}
\end{figure*}

Before using the data from the sensors, a preprocessing stage consisting of face detection, alignment, and normalization is performed. MTCNN algorithm \cite{zhang2016joint} was used for face detection in the color channel followed by face landmark detection using Supervised Descent Method (SDM) \cite{xiong2013supervised}. After these stages, the face image is aligned and converted to gray-scale with a resolution of $128 \times 128$ pixels. Since all the channels are aligned, these face locations are utilized for the alignment of other non-RGB channels as well. Also, normalization using  Mean Absolute Deviation (MAD) \cite{leys2013detecting} is performed to convert the raw 16-bit values to the 8-bit range. An example image after preprocessing stage is shown in Fig. \ref{fig:preprocessed}.
\subsubsection{Network architecture and training}

Since the data used is multi-channel, we use a multi-channel PAD framework called  `Multi-Channel Convolutional Neural Network'(\textit{MCCNN}) proposed in \cite{george_mccnn_tifs2019} as our base network. The main idea in \textit{MCCNN} was to use the joint representation from multiple channels for PAD task, leveraging a pretrained face recognition network. The \textit{MCCNN} architecture constituted of an extended version of LightCNN model \cite{wu2018light} adapted specifically for multi-channel PAD task. A pretrained LightCNN face recognition model was extended to accept multiple channels, and the embeddings from all channels were concatenated, and two fully connected layers were added on top of this joint representation layer for PAD task. The advantage in this architecture is that only lower layer features (which are known as Domain Specific Units (DSU) \cite{freitas2018heterogeneous} ) and higher-level fully connected layers are adapted in the training phase. The first fully connected layer contains ten nodes, and the second layer contains only one output node. The higher-level features in the LightCNN part are shared among all the modalities. This approach has two main advantages; first, there is a smaller number of parameters since the high-level features are shared across modalities, second, adapting only DSUs and final fully connected layers reduce possible over-fitting since PAD databases are typically small in size. An optimal set of layers to be adapted was obtained empirically and was used in the baseline \textit{MCCNN} and the proposed approach.

In our proposed approach, we use the same \textit{MCCNN} architecture, and the output from the penultimate fully connected layer was used as the embeddings. To quantify the effectiveness of our approach, we perform experiments on the \textit{MCCNN} architecture, while using both embeddings and the final output for the loss computation. An illustration of the proposed framework is shown in Fig. \ref{fig:arch}. At the time of training, both losses are used, and the model corresponding to the lowest validation score is selected. It is to be noted that, at the time of CNN training, both \textit{bonafide} and (known) attack samples are used.  After the CNN training, the network weights are frozen, and the \textit{bonafide} samples are feed-forwarded to obtain the embeddings.
\subsubsection{One-Class Gaussian Mixture Model}

After the training of \textit{MCCNN} with \textit{BCE} and \textit{OCCL}, the trained weights of the network are frozen, and it is used as a fixed feature extractor for the PAD task. Now that a compact representation is available, the objective is to learn a one-class classifier using the features obtained. We use  One-Class Gaussian Mixture Model for this task. The one class GMM is a generative approach which is used for modeling the distribution of the \textit{bonafide} class in the proposed framework.

A Gaussian Mixture Model is defined as the weighted sum of $K$ multivariate Gaussian distributions as:

\begin{equation}
\label{eq:gmm}
p(x|\Theta) = \sum_{k=1}^K w_k \mathcal{N}(x; \mu_k, \Sigma_k),
\end{equation}

where $\Theta = \{ w_k, \mu_k, \sigma_k \}_{ \{ k = 1,\dots,K \} }$ are the weights, means and the covariance matrix of the GMM.

Expectation-Maximization (EM) \cite{Dempster77maximumlikelihood} was used to compute the parameters of the GMM. A full covariance matrix is computed for each component, and the number of components to use was empirically selected as five ($K=5$).

During the training phase, embeddings obtained from \textit{bonafide} class only are used to train the One-Class GMM.

In test time, a sample is first forwarded though the network to obtain the embedding $x$, and then fed to the One-Class GMM to obtain the log-likelihood score as follows:

\begin{equation}
\label{eq:score}
score = log(p(x|\Theta))
\end{equation}

\begin{algorithm}[h]
\label{alg:algo_framework}
\SetAlgoLined
\DontPrintSemicolon

\KwData{($x_i$, $y_i$), where $x_i$ is multi-channel input and $y_i \in {0,1}$; 0 -- for attack and 1-- for \textit{bonafide}}
\KwResult{$W_C$ -- CNN weights, $\Theta_{GMM}$ -- Parameters of GMM}
\textbf{Constants} : $\lambda$ -- weighting factor, $\mu$ -- learning rate \;
\textbf{Initialize} : $C_{BF}$ -- center of \textit{bonafide} class, $W_C$ -- initial weights of CNN from pretrained model \

\For {mini-batch $\leftarrow$ 1 \textbf{to} P} {

Forward $x_i$ through the CNN \;
Compute the combined loss: $\mathcal{L} = (1-\lambda)\mathcal{L}_{BCE} + \lambda \mathcal{L}_{OCCL}$ \;

Back-propagate the loss and update the weights of DSUs and FC layers \;

Update the \textit{bonafide} center: \\ $c_{BF}= \hat{c}_{BF} (1-\alpha) + \alpha \frac{1}{N}\sum_{i=1}^{N}e_{i}$ \;

}

Forward $x_j$ (\textit{bonafide}, where $y_j=1$) through the CNN to obtain Embeddings $E_j$ \;

Estimate parameters of GMM from ${E_j}$:\\ $\Theta_{GMM}$= $(w_k,\mu_k, \Sigma_k)$ \;

\textbf{Parameters}$\leftarrow$ $(W_C,\Theta_{GMM})$\;

\caption{Algorithm for training the proposed framework}
\end{algorithm}

In summary, the proposed framework can be considered as a one-class classifier based framework for PAD. The crucial distinction is that, the features used are \textit{\textbf{learned}}. The loss function proposed forces the CNN to learn a compact representation for the \textit{bonafide} class leveraging the information from known attack classes. The algorithm for training the framework is shown in Algorithm \ref{alg:algo_framework}.

\subsection{Implementation details}

To increase the number of samples, data augmentation using random horizontal flips with a probability of 0.5 was used in training. Adam Optimizer \cite{kingma2014adam} was used to minimize the combined loss function. Learning rate of $1\times10^{-4}$ and a weight decay parameter of  $1\times10^{-5}$ was used. The network was trained for 50 epochs on GPU grid with a batch size of 32. The model corresponding to minimum validation loss in the $dev$ set is selected as the best model. For the four-channel models, the MCCNN architecture has about 13.1M parameters and about 14.5 GFLOPS. The implementation was done using PyTorch \cite{paszke2017automatic} library.

\section{Experiments}

In order to evaluate the effectiveness of the proposed approach, we have performed experiments in three publicly available databases, namely \textit{WMCA} \cite{george_mccnn_tifs2019}, \textit{MLFP} \cite{agarwal2017face}, and \textit{SiW-M} \cite{Liu_2019_CVPR} datasets.
Recently published \textit{CASIA-SURF} \cite{zhang2018casia} database also consists of multi-channel data, namely color, depth, and infrared channels with a limited set of attack instruments. However, the raw data from the sensors were not publicly available; in the publicly available version of the database, images were masked and scaled with custom preprocessing reducing the dynamic range of depth and infrared channels severely. Moreover, there was no guaranteed alignment between the channels. Therefore we can't use our framework with \textit{CASIA-SURF} database due to the mentioned limitations.

\subsection{WMCA dataset}

\begin{figure}[h]
     \centering
         \includegraphics[width=1\linewidth]{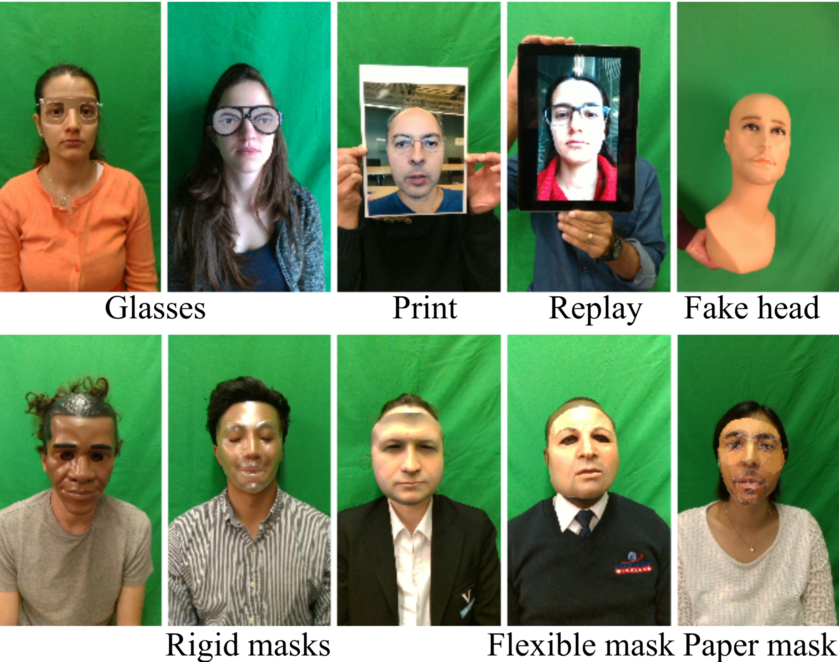}

\caption{Attack categories in \textit{WMCA} dataset, only RGB images are shown. Print and Replay constitutes the 2D attacks and all others are 3D attacks (Image taken from \cite{george_mccnn_tifs2019}).}
\label{fig:pa_wmca}
\end{figure}

\begin{figure}[h]
     \centering
         \includegraphics[width=0.7\linewidth]{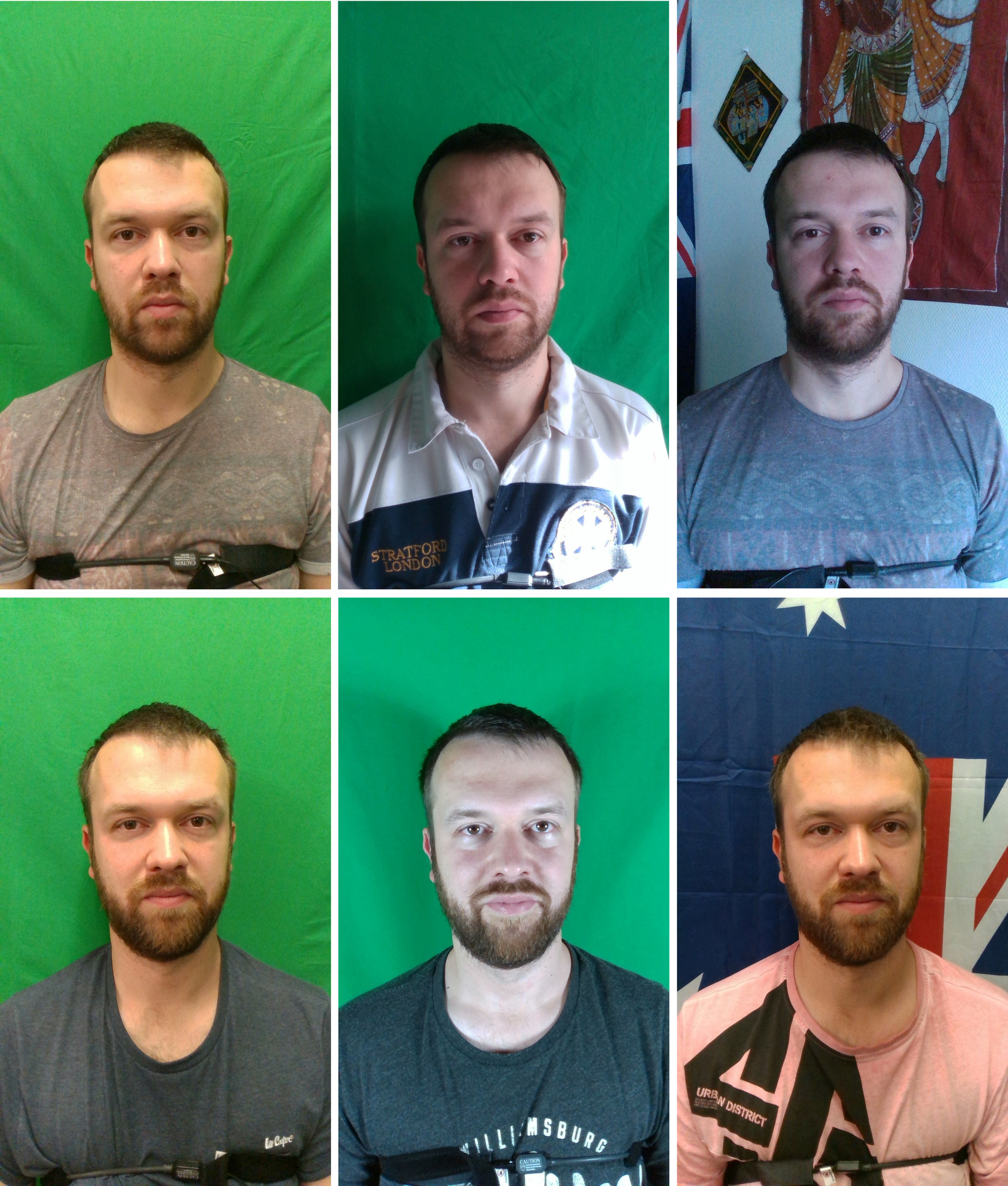}

\caption{Different sessions in \textit{WMCA} dataset, only RGB images are shown. A total of six sessions was used the \textit{WMCA} (Image taken from \cite{george_mccnn_tifs2019})}
\label{fig:sessions_wmca}
\end{figure}

 We have conducted an extensive set of experiments on \textit{Wide Multi-Channel presentation Attack} (\textit{WMCA}) \footnote{Database available at : \url{https://www.idiap.ch/dataset/wmca}} database, which contains a total of \textit{1679} video samples of \textit{bonafide} and attack attempts from \textit{72} identities. The database contains information from four different channels collected simultaneously, namely, color, depth, infrared, and thermal channels. The data was collected using two consumer devices,  Intel\textsuperscript{\textregistered} RealSense\texttrademark SR300 capturing RGB-NIR-Depth streams, and Seek Thermal CompactPRO for the thermal channel. The database contained around eighty different PAIs constituting seven different categories of attacks: print, replay, funny eyeglasses, fake head, rigid mask, flexible silicone mask, and paper masks. The RGB visualization of the attack categories is shown in Fig. \ref{fig:pa_wmca} and the different sessions in Fig. \ref{fig:sessions_wmca}. Detailed information about the \textit{WMCA} database can be found in the publication \cite{george_mccnn_tifs2019}. The statistics of the number of samples in each category and their types are shown in Table \ref{tab:stats_wmca}. We have made challenging protocols in the \textit{WMCA} dataset to perform an extensive set of evaluations emulating real-world unseen attack scenarios.


\subsubsection{Protocols in \textit{WMCA}}

To test the performance of the algorithm in known and unseen attack scenarios, we created three protocols in the \textit{WMCA} dataset. The protocols are described below.

\begin{itemize}
    \item \textbf{grandtest} : This is the exact same \textit{grandtest} protocol available with \textit{WMCA} database, here all the attack types are present in almost equal proportions in the \textit{train}, \textit{development} and \textit{evaluation} sets. The attack types and \textit{bonafide} samples are divided into three folds, and the client ids are disjoint across the three sets. Each presentation attack instrument had a separate client id. The train, dev, eval splits were made in such a way that a specific PA instrument will appear in only one fold.
    \item \textbf{unseen-2D} : In this protocol, we use same splits as \textit{grandtest} and removed all 2D attacks from \textit{train} and \textit{development} groups. \textit{Evaluation} set contains only \textit{bonafide} and 2D attacks. This emulates the performance of a system when encountered with 2D attacks which was not seen in training.
    \item \textbf{unseen-3D} : In this protocol, we use same splits as \textit{grandtest} and removed all 3D attacks from \textit{train} and \textit{development} groups. \textit{Evaluation} set contains only \textit{bonafide} and 3D attacks. This emulates the performance of a system when encountered with 3D attacks which were not seen in training. This is the most challenging protocol as the model sees only the simpler 2D attacks in training and encounter challenging 3D attacks in testing.
\end{itemize}

While the \textit{grandtest} protocol emulates the known attack scenario, other protocols emulate the unseen attack scenario. All protocols are made available publicly.

\begin{table}[ht]
\centering
\caption{Statistics of attacks in \textit{WMCA} database}
\label{tab:stats_wmca}
\resizebox{0.35\textwidth}{!}{%

\begin{tabular}{ccc}
\toprule
PA Category                        & Type & \#Presentations                \\ \midrule
\rowcolor{Gray}

\textit{bonafide} & -    & 347                            \\
glasses                            & Partial   & 75                             \\
\rowcolor{Gray}
print                              & 2D   & 200                            \\
replay                             & 2D   & 348                            \\
\rowcolor{Gray}
fake head                          & 3D   & 122                            \\
rigid mask                         & 3D   & 137                            \\
\rowcolor{Gray}

flexible mask                      & 3D   & 379                            \\
paper mask                         & 3D   & 71                             \\ \midrule
\rowcolor{Gray}

\textbf{TOTAL}    &      & \textbf{1679} \\ \bottomrule
\end{tabular}
}
\end{table}

\subsection{MLFP dataset}

MLFP dataset \cite{agarwal2017face} consists of attacks captured with seven 3D latex masks and three 2D print attacks. The dataset contains videos captured from color, thermal and infrared channels. Since channels were captured individually in different recording sessions, multi-channel approaches are not trivial. Also, the alignment of channels is not possible since they are not collected simultaneously. Hence, we only use the RGB videos from the MLFP dataset for our experiments. The database contains videos of 10 subjects wearing both print and latex masks. There are 440 videos are consisting of both attacks and \textit{bonafide} for the RGB channel.

\subsubsection{Protocols in \textit{MLFP}}

To emulate known and unseen attack scenarios, we created three new protocols in the \textit{MLFP} dataset. There are two types of attacks, namely print and mask. Only two sets, i.e., \textit{train} and \textit{evaluation} are created due to the small size of the dataset. We used a subset of the train set (10\%) for model selection. The protocols are described below.

\begin{itemize}
    \item \textbf{grandtest} : This protocol emulates the known attack scenario. Both the attacks are present in both \textit{train} and \textit{evaluation} set. However, the subjects and the PAs are disjoint across the two sets.
    \item \textbf{unseen-print} : In this protocol, only \textit{bonafide} and mask attacks are present in \textit{train} set; the \textit{evaluation} set contains only \textit{bonafide} and print attacks. This emulates unseen attack scenario.
    \item \textbf{unseen-mask} : In this protocol, only \textit{bonafide} and print attacks are present in \textit{train} set; the \textit{evaluation} set contains only \textit{bonafide} and mask attacks. This protocol also emulates unseen attack scenario.
\end{itemize}

\subsection{SiW-M dataset}

The Spoof in the Wild database with Multiple Attack Types (\textit{SiW-M}) \cite{Liu_2019_CVPR} consists of a wide variety of attacks captured only in RGB spectra. The database consists of images from 493 subjects, and a total of 660 \textit{bonafide} and 968 attack samples. A total of 1628 files, consisting of 13 different attack types, collected in different sessions, pose, lighting, and expression (PIE) variations. The attacks consist of various types of masks, makeups, partial attacks, and 2D attacks. The videos are available in 1080P resolution.

\subsubsection{Protocols in \textit{SiW-M}}

To emulate unseen attack scenarios, we use the leave-one-out (LOO) testing protocols available with the \textit{SiW-M} \cite{Liu_2019_CVPR} dataset. The protocols consists of only \textit{train} and \textit{eval} sets. In each LOO protocol, the training set consists of 80\% percentage of the live data and 12 types of spoof attacks. The evaluation set consists of 20\% of \textit{bonafide} data and the attack which was left out in the training phase. The subjects in \textit{bonafide} sets are disjoint in \textit{train} and \textit{evaluation} sets. A subset of the train set (5\%) was used for model selection.  Additionally, we have created a \textit{grandtest} protocol, specifically for cross-database testing which contains all the attack types in all the folds.

\subsection{Evaluation metrics}

We report the standardized ISO/IEC 30107-3 metrics \cite{ISO}, Attack Presentation Classification Error Rate (APCER),  and Bonafide Presentation Classification Error Rate (BPCER), and  Average Classification Error Rate (ACER) in the  $test$ set. A BPCER threshold of 1\% is used for computing the threshold in $dev$ set. The APCER and BPCER in both \textit{dev} and \textit{eval} sets are also reported. Additionally, the ROC curves for experiments are also shown in all the protocols. For the \textit{MLFP} dataset, we report only EER in the \textit{evaluation} set since only two sets are available. For SiW-M database, we apply a threshold selected a-priori in all protocols, for computing the metrics, to be comparable with the results in \cite{Liu_2019_CVPR}.
\subsection{Baselines}

We have implemented three feature-based baselines and two CNN based baselines. For a fair comparison, all the benchmarks are multi-channel methods and use the same four channels. Besides, an RGB only CNN model is also added for comparison.
A short description of the baselines along with the acronyms used are shown below:

\begin{itemize}

\item \textit{MC-RDWT-Haralick-SVM}: This baseline is the multi-channel extension of the RDWT-Haralick-SVM approach proposed in \cite{agarwal2017face}; the images from all channels are stacked together after preprocessing. For each channel, the image is divided into a  $4 \times 4$ grid, and Haralick \cite{haralick1979statistical} features obtained from the RDWT decompositions are concatenated from all the grids in all channels to get the joint feature vector. The joint feature is used with a linear SVM for PAD.

\item \textit{MC-RDWT-Haralick-GMM}: Here, the feature extraction stage is same as \textit{MC-RDWT-Haralick-SVM}; however, the classifier used is one class GMM. Only \textit{bonafide} samples are used in training this model. This model is added to show the performance of one class models in unseen attack scenarios.

\item \textit{MC-LBP-SVM}: Here, again, the same preprocessing is performed on all the channels first. After this, Spatially enhanced histograms of LBP representation from all the component channels are computed and concatenated to a feature vector. The features extracted are fed to an SVM for PAD task.

\item \textit{DeepPixBiS} :  This is a CNN based system \cite{george2019deep} trained using both binary and pixel-wise binary loss function. This model only uses RGB information for PAD.

\item \textit{MC-ResNetPAD}: We reimplemented the architecture from \cite{parkin2019recognizing} extending it to four channels, based on their open-source implementation \footnote{Available from: \url{https://github.com/AlexanderParkin/ChaLearn_liveness_challenge}}. This approach obtained the first place solution in the `CASIA-SURF' challenge. For a fair comparison, instead of using an ensemble we used the best pretrained model as suggested in \cite{parkin2019recognizing}.

\item \textit{MCCNN(BCE)} : This is the multi-channel CNN system described in \cite{george_mccnn_tifs2019}, which achieved state of the art performance in the \textit{grandtest} protocol. The model is trained using Binary Cross-Entropy (\textit{BCE}) loss only.
\end{itemize}

All the baseline methods described are reproducible, and the details about the parameters can be found in our open-source package \footnote{Source code: \url{https://gitlab.idiap.ch/bob/bob.paper.oneclass_mccnn_2019}}.

\subsection{Experiments and Results in \textit{WMCA} dataset}

\begin{figure*}[h!]
\centering
  \subfloat[grandtest]{\includegraphics[height=5.2cm,page=2]{figures/grandtest_wmca.pdf}}%
\hfil
  \subfloat[unseen-2D]{\includegraphics[height=5.2cm,page=2]{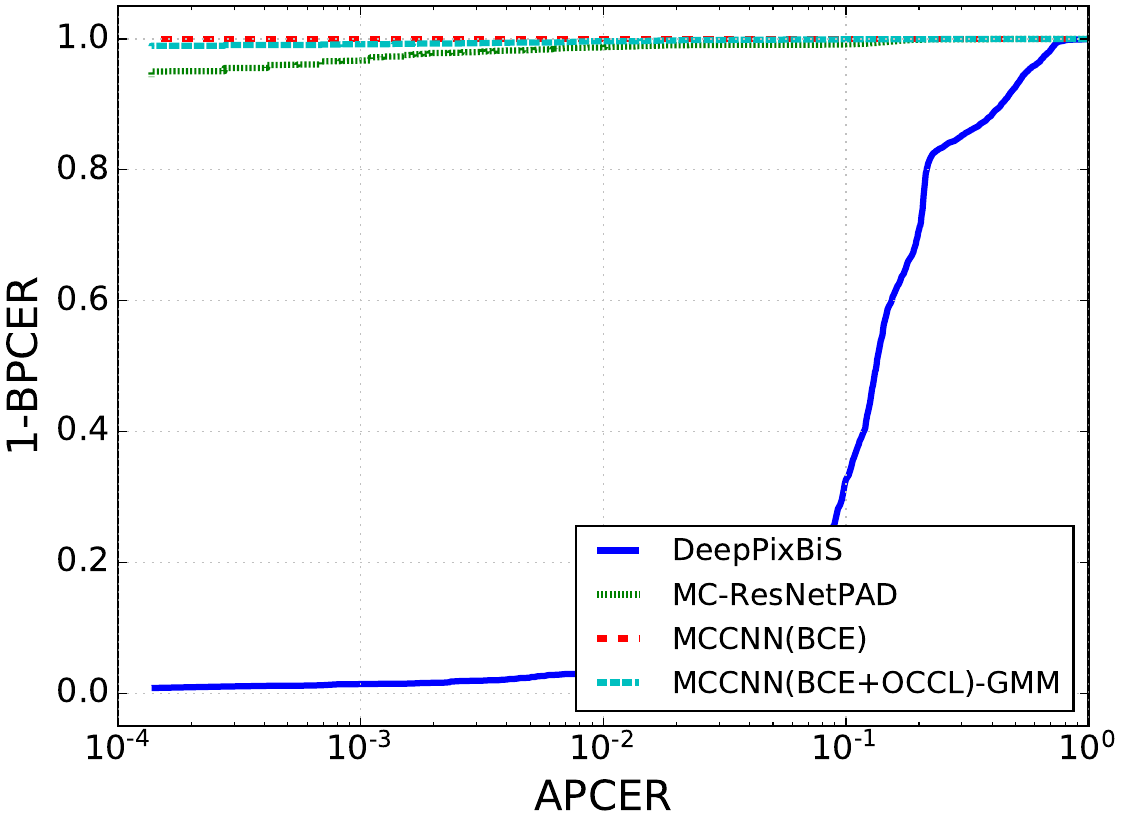}}%
\hfil
  \subfloat[unseen-3D]{\includegraphics[height=5.2cm,page=2]{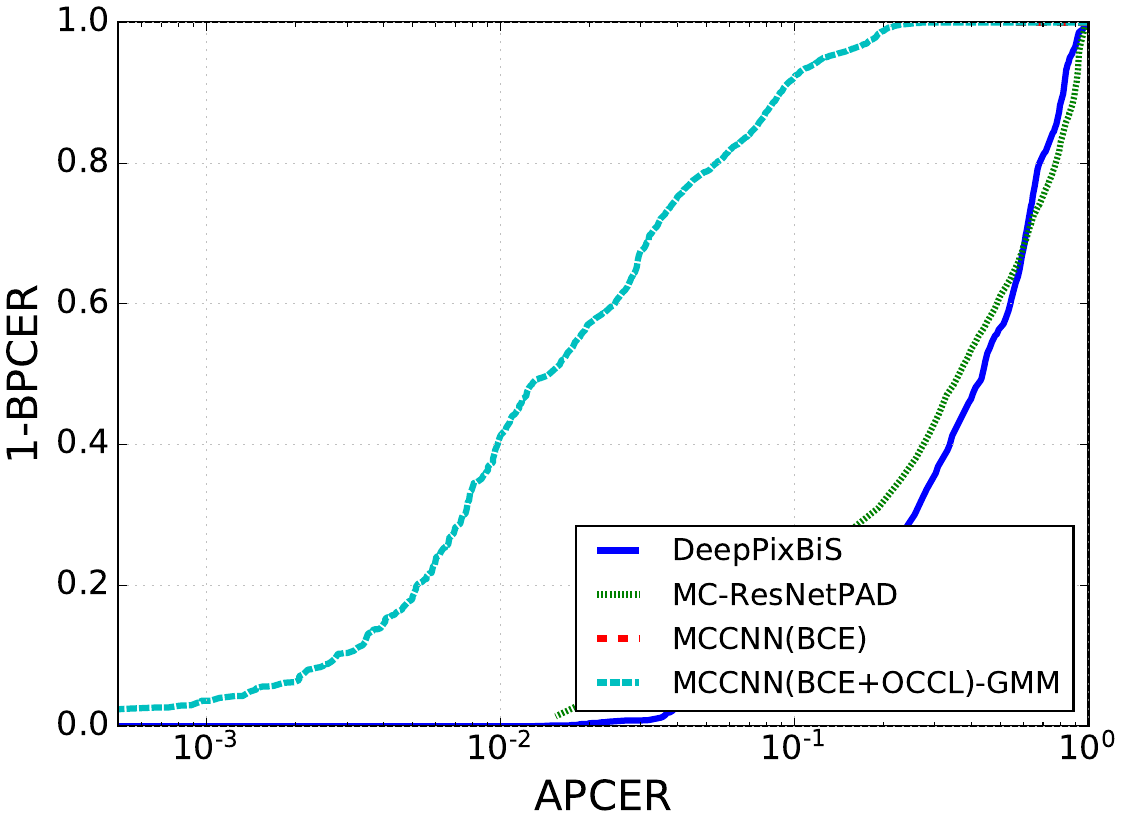}}%
\caption{DET curves for the \textit{eval} sets of different protocols of \textit{WMCA} dataset a) \textit{grandtest}, b) \textit{unseen-2D}, c) \textit{unseen-3D} protocol.}
 \label{fig:ROC}
\end{figure*}

\begin{table*}[h!]
\centering
\caption{Performance of the baseline systems and the proposed method in \textit{grandtest} protocol of \textit{WMCA} dataset. The values reported are obtained with a threshold computed for BPCER 1\% in $dev$ set.}
\label{tab:grandtest}
\resizebox{0.6\textwidth}{!}{%
{

\begin{tabular}{@{}ccc|ccc@{}}
\toprule
\multirow{2}{*}{Method} & \multicolumn{2}{c|}{dev (\%)}     & \multicolumn{3}{c}{test (\%)}                                  \\ \cmidrule(r){2-6}
                        & \multicolumn{1}{c|}{APCER} & ACER & \multicolumn{1}{c|}{APCER} & \multicolumn{1}{c|}{BPCER} & ACER  \\ \midrule
\rowcolor{Gray}
MC-RDWT-Haralick-SVM & 3.6  &2.3  &5.4  &1.2  &3.3 \\
MC-LBP-SVM  & 3.6  &2.3  &8.5  &0.6  &4.6 \\
\rowcolor{Gray}
MC-RDWT-Haralick-GMM  &43.4  &22.2  &47.7  &1.7  &24.7 \\
DeepPixBiS (RGB only)\cite{george2019deep} &1.0  &1.0   & 8.2   &3.7 &6 \\
\rowcolor{Gray}
MC-ResNetPAD \cite{parkin2019recognizing} &3.8 &2.4 &3.5 &1.6 &2.6 \\
MCCNN(BCE)\cite{george_mccnn_tifs2019} & 0.4 & 0.7 &0.5 &0 & \textbf{0.2}\\  \midrule
\rowcolor{Gray}
\textbf{MCCNN(BCE+OCCL)-GMM} &0.1  &0.6  &0.6  &0.1  &0.4 \\
 \bottomrule
\end{tabular}
}

}
\end{table*}

\begin{table*}[h!]

\centering
\caption{Performance of the baseline systems and the proposed method in \textbf{unseen} protocols of \textit{WMCA} dataset. The values reported are obtained with a threshold computed for BPCER 1\% in $dev$ set.}
\label{tab:unseen}

\resizebox{0.65\textwidth}{!}{%
{

\begin{tabular}{@{}cccc|ccc@{}}
\toprule
\multirow{2}{*}{Method}                            & \multicolumn{3}{c|}{unseen-2D}                                                       & \multicolumn{3}{c}{unseen-3D}                                 \\ \cmidrule(r){2-7}
                                                   & \multicolumn{1}{c|}{APCER} & \multicolumn{1}{c|}{BPCER} & \multicolumn{1}{c|}{ACER}  & \multicolumn{1}{c|}{APCER} & \multicolumn{1}{c|}{BPCER} & ACER \\ \midrule
\rowcolor{Gray}

MC-RDWT-Haralick-SVM                               & 0.3        &0.1                    & \multicolumn{1}{c|}{0.2}  &  66.0           &0.1                        & 33.1    \\
MC-LBP-SVM                                         & 40.7       & 0.1                   & \multicolumn{1}{c|}{20.4} & 38.9            &0.2                        & 19.5    \\
\rowcolor{Gray}

MC-RDWT-Haralick-GMM                               & 0.0         &0.2                    & \multicolumn{1}{c|}{\textbf{0.1}}  & 70.8           &1.9                        &36.4     \\
DeepPixBiS (RGB only)\cite{george2019deep}                              & 77.7           &0.3                    & \multicolumn{1}{c|}{39} & 74.7          & 16.3                       &45.5      \\
\rowcolor{Gray}
MC-ResNetPAD \cite{parkin2019recognizing}          &4.1           &0.9                  & \multicolumn{1}{c|}{2.5}    & 92.2          &6.4	                       &49.3 \\
MCCNN(BCE)\cite{george_mccnn_tifs2019}             & 0.0        &1.0                   & \multicolumn{1}{c|}{0.5}  & 62.0                     & 0.0                     & 31.0    \\ \midrule
\rowcolor{Gray}
\textbf{MCCNN(BCE+OCCL)-GMM}                     &0.3             &0.6                       & \multicolumn{1}{c|}{0.5}      &15.4                           &3.9                           &\textbf{9.7}      \\
\bottomrule
\end{tabular}

}
}
\end{table*}

\begin{figure*}[ht]

\centering
  \subfloat[grandtest]{\includegraphics[height=5.2cm]{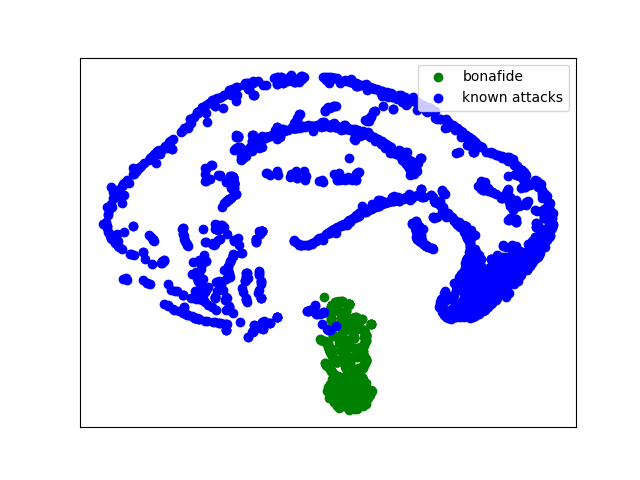}}%
\hspace{-4.2em}
  \subfloat[unseen-2D]{\includegraphics[height=5.2cm]{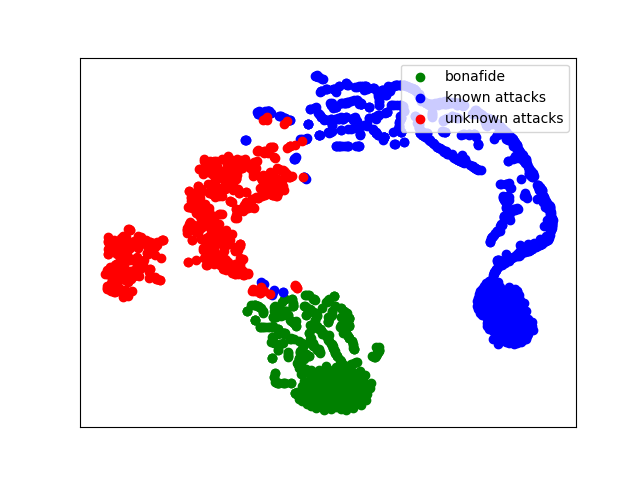}}%
\hspace{-4.2em}
  \subfloat[unseen-3D]{\includegraphics[height=5.2cm]{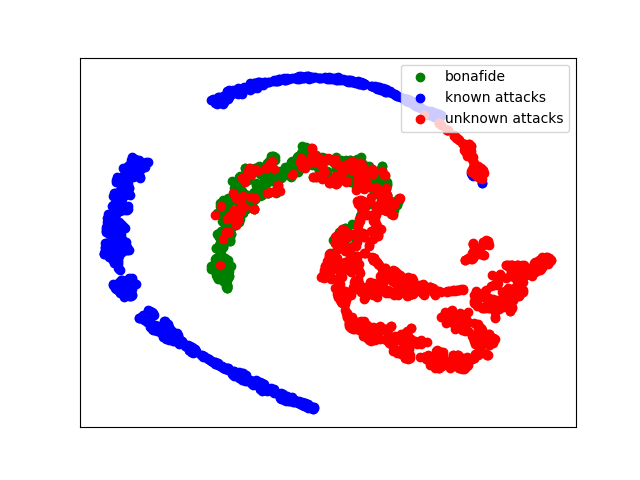}}%
\vspace{-2em}

 \centering
  \subfloat[grandtest]{\includegraphics[height=5.2cm]{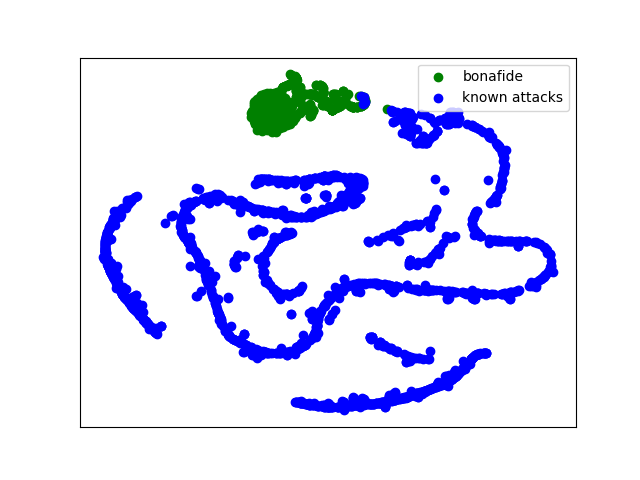}}%
\hspace{-4.2em}
  \subfloat[unseen-2D]{\includegraphics[height=5.2cm]{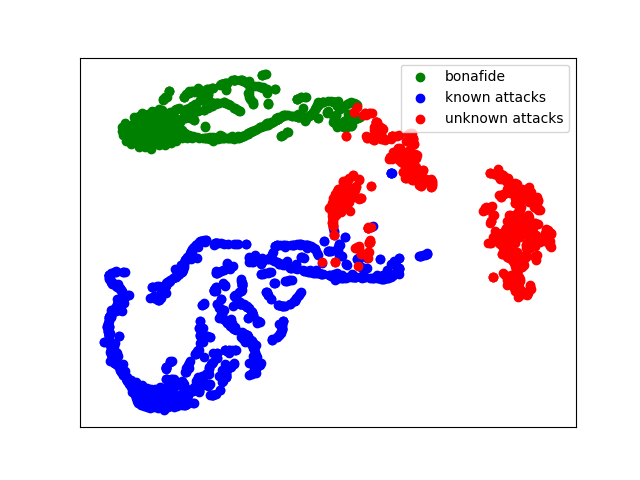}}%
\hspace{-4.2em}
  \subfloat[unseen-3D]{\includegraphics[height=5.2cm]{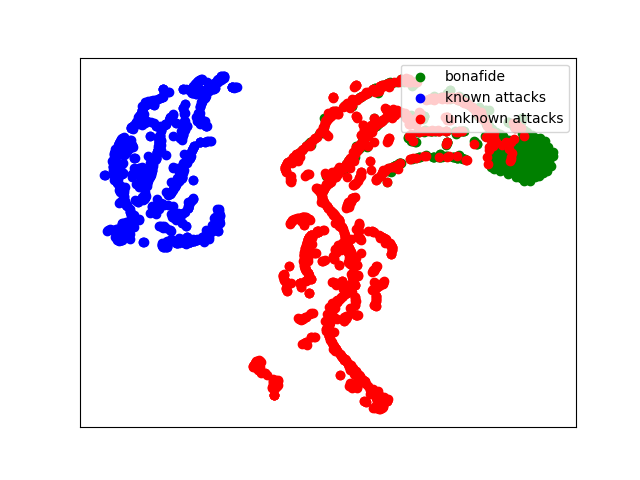}}%
\caption{t-SNE plots of embeddings in the protocols in \textit{WMCA} dataset. First row (a,b,c) shows the embeddings when only \textit{BCE} loss was used. Second row (d,e,f) shows the embeddings when both the losses are used. Embeddings of both known and unseen attacks are shown in the figures for each protocol. Grandtest protocol contains only known attacks in the test set.}
 \label{fig:tsne}

\end{figure*}

\begin{figure}[h]
     \centering
         \includegraphics[width=0.65\linewidth]{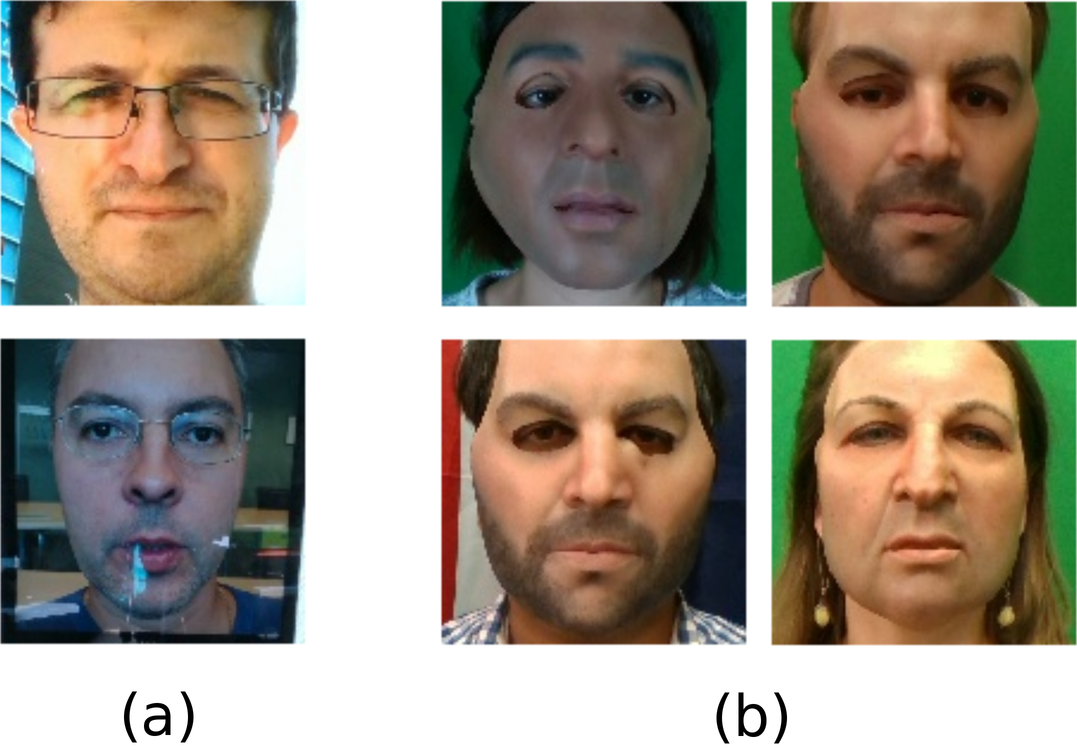}

\caption{The attack samples which are closer to \textit{bonafide} cluster in a) unseen-2D (Fig.8(E)) and b) unseen-3D ((Fig.8(F))) protocol for the proposed framework.}
 \label{fig:tsne_close}
\end{figure}

We have tested the baselines and the proposed approach in three different protocols in \textit{WMCA}. The proposed approach is denoted as \textit{MCCNN(BCE+OCCL)-GMM}.

\begin{itemize}

\item \textit{MCCNN(BCE+OCCL)-GMM}: Here, the \textit{bonafide} embeddings from the \textit{MCCNN} trained using both the losses are used to train a GMM, and in the evaluation stage, the score from the one class GMM is used as the PAD score.

\end{itemize}

The results in each protocol are described below.
\subsubsection{Experiments in \textit{grandtest} protocol}

The \textit{grandtest} protocol emulates the known attack scenario. Table \ref{tab:grandtest} tabulates the results in the \textit{grandtest} protocol. The proposed approach outperforms the feature-based methods by a large margin as expected. The model \textit{MC-RDWT-Haralick-GMM} trained using a one-class model achieves the worse results. It is interesting to note that the \textit{MC-RDWT-Haralick-SVM} model, trained using the same feature as a binary classifier performed much better. This shows one weakness of one-class classifiers in a known attack scenario, as they do not use the known attacks in training.  The \textit{MCCNN(BCE)} achieves much better performance as compared to \textit{MC-ResNetPAD}. The \textit{MCCNN(BCE)} trained as a binary classifier achieves the best performance in this protocol. The proposed \textit{MCCNN(BCE+OCCL)-GMM} approach achieves comparable performance to \textit{MCCNN(BCE)}. This indicates that the one class GMM classifier performs on par with the binary classification, provided they are trained with compact feature representations.

\subsubsection{Experiments in \textit{unseen-2D} and \textit{unseen-3D} protocol}

The \textit{unseen-2D} and \textit{unseen-3D} protocols emulates the unseen attack scenario. The \textit{unseen-3D}  is the most challenging protocol since it is trained only on 2D - print and replay attacks and encounters a wide variety of 3D attacks such as silicone masks, fake heads, mannequins, etc. in the \textit{eval} set.

Most of the approaches perform well in the \textit{unseen-2D} protocol. This result is intuitive as these models are trained on challenging 3D attacks, detection of 2D attacks is much easier. Moreover, the 2D attacks can be easily identified in depth, thermal, and infrared channels. Even some feature-based methods perform well in this protocol, with \textit{MC-RDWT-Haralick-GMM} method achieving the best performance. This shows the advantage of one class model in an unseen attack scenario. The proposed approach \textit{MCCNN(BCE+OCCL)-GMM} and \textit{MCCNN(BCE)} baseline perform comparably in this protocol. Notably, the DeepPixBiS model achieves much worse results in this protocol. This could be because discriminating between \textit{bonafide} and 2D attacks are harder when only RGB information is used.

The \textit{unseen-3D} protocol shows important results. All the baselines show inferior performance when encountered with unseen 3D samples. This shows the failure of binary classifiers in generalizing to challenging unseen attacks. The \textit{MCCNN(BCE)} approach, while being architecturally similar, fails to generalize when trained in the binary classification setting. With the proposed approach, performance improves to 9.7\% when the one class GMM is used on the \textit{bonafide} representations. Since the network learns to map the \textit{bonafide} samples to a compact cluster in the feature space, even in the presence of unseen attacks, the decision boundary learned for the \textit{bonafide} class is robust. The unseen attacks map far from the \textit{bonafide} cluster and hence becomes easy to discriminate from \textit{bonafide} samples. This result is encouraging since the network was shown only 2D attacks in training, and still, it manages to achieve good performance against challenging 3D attacks. The ROCs for all the protocols are shown in Fig. \ref{fig:ROC}.

The t-SNE \cite{maaten2008visualizing} plots of the embeddings for all protocols are shown in Fig. \ref{fig:tsne}. Five frames from each video in the evaluation sets of the protocols are used for this visualization. While the difference between \textit{bonafide} and attacks are clear in the \textit{grandtest} and \textit{unseen-2D}, difference in \textit{unseen-3D} protocol is very evident. It can be clearly seen that the \textit{bonafide} class clusters together and is far from the \textit{bonafide} representation in the embedding space in the \textit{unseen-3D} protocol when the proposed loss is used. Unseen attacks overlaps with \textit{bonafide} embeddings when only \textit{BCE} is used. This clearly demonstrates the effectiveness of the proposed approach for unseen attack detection. The unseen attacks which are overlapping with the \textit{bonafide} region are shown in Fig. \ref{fig:tsne_close}. It can be seen that some video replay samples and flexible silicone 3D masks get misclassified in unseen-2D and unseen-3D protocols respectively.

\subsubsection{Ablation study with channels}

\begin{figure*}[ht!]
\centering
  \subfloat[grandtest]{\includegraphics[height=5.2cm,page=2]{figures/ablation_grandtest.pdf}}%
\hfil
  \subfloat[unseen-2D]{\includegraphics[height=5.2cm,page=2]{figures/ablation_u2d.pdf}}%
\hfil
  \subfloat[unseen-3D]{\includegraphics[height=5.2cm,page=2]{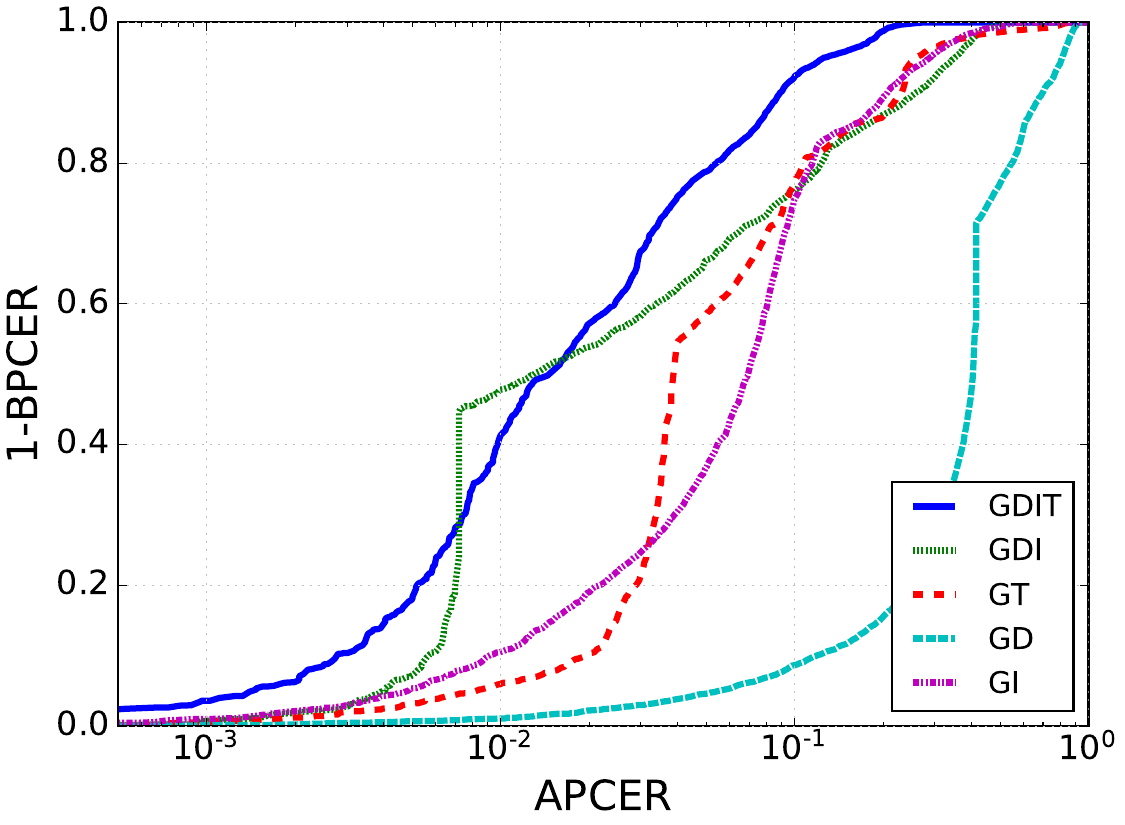}}%
\caption{Ablation study with different combination of channels, DET curves for the \textit{eval} sets of different protocols of \textit{WMCA} dataset a) \textit{grandtest}, b) \textit{unseen-2D}, c) \textit{unseen-3D} protocol.}
 \label{fig:ablation_ROC}
\end{figure*}

To evaluate the performance of the proposed framework on different set of channels, we perform an ablation study by including a different set of channels. We used only the best performing \textit{MCCNN(BCE+OCCL)-GMM} approach in this ablation study. In all combinations, the gray-scale channel is present since it is used as a reference. This is required as the embedding from the gray-scale part can be used for face recognition as well.

The acronyms for different channels are shown below:

\begin{itemize}
\item G: Gray-scale image
\item D: Depth image
\item I: Infrared channel
\item T: Thermal channel
\end{itemize}

Various combinations of these channels are experimented with, and the results are tabulated in Table \ref{tab:ablation_channels}. It is to be noted
that the channels G, D and I come from the same device and T is coming from a different device. Usually, thermal cameras are expensive, compared to RGB-D cameras, and hence the combinations involving subsets of G, D and I are more interesting from a deployment point of view.

\begin{table}[h!]

\centering
\caption{ Performance of the proposed framework with different combinations of channels in all protocols of \textit{WMCA} dataset. The values reported are obtained with a threshold computed for BPCER 1\% in $dev$ set.}
\label{tab:ablation_channels}

\resizebox{0.4\textwidth}{!}{%
{

\begin{tabular}{cccc}
\toprule

\multirow{2}{*}{Channels} & \multicolumn{1}{c|}{grandtest} & \multicolumn{1}{c|}{unseen-2D} & unseen-3D                     \\ \cline{2-4}
                          & \multicolumn{1}{c|}{ACER}      & \multicolumn{1}{c|}{ACER}       & ACER                           \\ \midrule
\rowcolor{Gray}

GDIT                      &\textbf{0.4}                 &\textbf{0.5}                    &\textbf{9.7}                  \\
GDI                       &1.1                          &11.2                            &23.1                  \\
\rowcolor{Gray}

GT                        &2.2                          &3.2                             &21.5                  \\
GD                        &2.3                          &49.4                            &45.4                 \\
\rowcolor{Gray}

GI                        &1.1                          &2.2                             &22.6                  \\ \bottomrule
\end{tabular}

}
}
\end{table}

From Table \ref{tab:ablation_channels}, it can be seen that the performance degrades as channels are removed. However, the combination GI achieves reasonable performance while considering the performance-cost ratio. The ROCs for different protocols are shown in Fig. \ref{fig:ablation_ROC}.

\subsection{Experiments and Results in \textit{MLFP} dataset}

We have used only the RGB channel for the experiments since the other channels were not captured simultaneously. For the MCCNN framework and other baselines, `R', `G', and `B' are considered as the different channels in these experiments.
We have performed the experiments in the three newly created protocols and the results are tabulated in Table \ref{tab:mlfp_results}.

From the results in Table \ref{tab:mlfp_results}, it can be seen that the CNN based approach outperforms the feature-based approaches. The MCCNN framework, with the
addition of the newly proposed loss outperforms the architecture trained with BCE only, showing the effectiveness of the proposed approach.

Even though the proposed approach performs better than the baselines, it is to be noted that the key point of the proposed approach, leveraging multi-channel information, is not utilized here. The architecture is not optimized for PAD in RGB and this experiment is performed only to show the change in performance with the new loss function. Nevertheless, the proposed approach achieves better performance as compared to the baselines in all the protocols.

\begin{table}[h]
\centering
\caption{Performance of the proposed framework in the protocols in \textit{MLFP} dataset. Only RGB channel was used in this experiments. The values reported are the EER in the \textit{evaluation} set.}
\label{tab:mlfp_results}
\resizebox{0.5\textwidth}{!}{%
{

\begin{tabular}{cccc}
\toprule
Algorithm                     &grandtest &\begin{tabular}[c]{@{}c@{}}unseen\\ print\end{tabular} &\begin{tabular}[c]{@{}c@{}}unseen\\ mask\end{tabular} \\ \midrule
\rowcolor{Gray}

MC-RDWT-Haralick-SVM               &9.8       &12.0       &32.2      \\
MC-LBP-SVM                    &6.3       &27.1       &9.3       \\
\rowcolor{Gray}
MC-RDWT-Haralick-GMM          & 27.4         &40.8           &21.5           \\
DeepPixBiS (RGB only)\cite{george2019deep}         & 6.3           &24.8                   &17.5 \\
\rowcolor{Gray}
MCCNN (BCE)                   &5.5       &9.2        &5.2       \\ \midrule
\textbf{MCCNN (BCE+OCCL)-GMM}  &\textbf{1.2}       &\textbf{3.3}        &\textbf{3.4}       \\ \bottomrule
\end{tabular}

}
}
\end{table}

\subsection{Experiments and Results in \textit{SiW-M} dataset}

\begin{table*}[ht!]
\small
	\centering
	\caption{Performance of the proposed framework in the leave one out protocols in \textit{SiW-M} dataset. Only RGB channel was present in this dataset.}
	\vspace{-3mm}
	\resizebox{\textwidth}{!}
{
{

	\begin{tabular}{l|l|c|c|c|c|c|c|c|c|c|c|c|c|c|c}
	\toprule
	\multirow{2}{*}{Methods} & \multirow{2}{*}{Metrics (\%)} &\multirow{2}{*}{Replay}& \multirow{2}{*}{Print} & \multicolumn{5}{c|}{Mask Attacks}  & \multicolumn{3}{c|}{Makeup Attacks}  &  \multicolumn{3}{c|}{Partial Attacks} & \multirow{2}{*}{Average}\\ \cline{5-15}
	 &&&  & Half & Silicone & Trans. & Paper & Manne. & Obfusc. & Imperson. & Cosmetic & Funny Eye & Paper Glasses & Partial Paper &\\ \midrule

    \multirow{4}{*}{MC-RDWT-Haralick-SVM }
	& APCER &  19.80 & 19.15 &  30.76 &  28.15 &   33.35 &   0.29 &    4.50 &  68.91 &     0.00 &   35.20 &  53.12 &    34.53 &    3.49 & $ 25.4 \pm 20.8 $ \\ \cline{3-16}
	& BPCER &  14.50 & 13.89 &  14.66 &  16.83 &   15.38 &  16.68 &   15.88 &  16.03 &    16.53 &   16.37 &  14.58 &    14.47 &   15.73 & $ 15.5 \pm  0.9 $ \\ \cline{3-16}
	& ACER  &  17.15 & 16.52 &  22.71 &  22.49 &   24.37 &   8.49 &   10.19 &  42.47 &     8.26 &   25.79 &  33.85 &    24.50 &    9.61 & $ 20.4 \pm 10.3 $ \\ \cline{3-16}
	& EER   &  16.88 & 16.53 &  21.80 &  20.73 &   21.94 &   7.34 &    9.88 &  32.56 &     2.37 &   23.51 &  31.72 &    21.94 &   10.05 & $ 18.2 \pm  9.0 $ \\ \midrule

	\multirow{4}{*}{MC-LBP-SVM }
	&APCER &  10.77 &  12.91 &  10.28 &  35.19 &  37.78 &   0.59 &   6.50 &  96.09 &   0.00 &  26.00 & 40.91 &  35.51 &   2.73 & $ 24.2 \pm 26.3 $ \\ \cline{3-16}
	&BPCER &  22.90 &  22.18 &  22.48 &  22.33 &  23.13 &  23.70 &  23.59 &  22.79 &  23.93 &  22.90 & 19.92 &  21.11 &  23.74 & $ 22.6 \pm  1.1 $ \\ \cline{3-16}
	&ACER  &  16.83 &  17.54 &  16.38 &  28.76 &  30.46 &  12.15 &  15.04 &  59.44 &  11.97 &  24.45 & 30.41 &  28.31 &  13.24 & $ 23.4 \pm 12.9 $ \\ \cline{3-16}
	&EER   &  15.96 &  16.83 &  16.87 &  28.51 &  29.77 &  10.54 &  12.75 &  52.60 &   1.90 &  24.61 & 28.32 &  26.76 &  11.29 & $ 21.2 \pm 12.6 $ \\ \midrule

    \multirow{4}{*}{Auxiliary \cite{liu2018learning} }
     & APCER &$23.7$& $7.3$ & $27.7$ & $18.2$ & $97.8$ & $8.3$ & $16.2$ & $100.0$ & $18.0$ & $16.3$ & $91.8$ & $72.2$ & $0.4$ & $38.3\pm37.4$  \\ \cline{3-16}
     & BPCER &$10.1$& $6.5$ & $10.9$ & $11.6$ & $6.2$ & $7.8$ & $9.3$ & $11.6$ & $9.3$ & $7.1$ & $6.2$ & $8.8$ & $10.3$ & $8.9\pm2.0$  \\ \cline{3-16}
     & ACER &$16.8$& $6.9$ & $19.3$ & $14.9$ & $52.1$ & $8.0$ & $12.8$ & $55.8$ & $13.7$ & $11.7$ & $49.0$ & $40.5$ & $5.3$ &$23.6\pm18.5$  \\ \cline{3-16}
     & EER &$14.0$& $4.3$ & $11.6$ & $12.4$ & $24.6$ & $7.8$ & $10.0$ & $72.3$ & $10.1$ & $9.4$ & $21.4$ & $18.6$ & $4.0$  & $17.0\pm17.7$ \\ \midrule

    \multirow{4}{*}{Deep Tree Network \cite{Liu_2019_CVPR}}

     & APCER &$1.0$& $0.0$ & $0.7$ & $24.5$ & $58.6$ & $0.5$ & $3.8$ & $73.2$ & $13.2$ & $12.4$ & $17.0$ & $17.0$ & $0.2$ & $17.1\pm23.3$  \\ \cline{3-16}
     & BPCER &$18.6$& $11.9$ & $29.3$ & $12.8$ & $13.4$ & $8.5$ & $23.0$ & $11.5$ & $9.6$ & $16.0$ & $21.5$ & $22.6$ & $16.8$ & $16.6\pm6.2$  \\ \cline{3-16}
     & ACER &$9.8$& $6.0$ & $15.0$ & $18.7$ & $36.0$ & $4.5$ & $7.7$ & $48.1$ & $11.4$ & $14.2$ & $19.3$ & $19.8$ & $8.5$ & $16.8\pm11.1$  \\ \cline{3-16}
     & EER &$10.0$& $2.1$ & $14.4$ & $18.6$ & $26.5$ & $5.7$ & $9.6$ & $50.2$ & $10.1$ & $13.2$ & $19.8$ & $20.5$ & $8.8$ & $16.1\pm12.2$   \\ \midrule

    \multirow{4}{*}{DeepPixBiS \cite{george2019deep}}

    & APCER & 19.18 & 8.97 &  1.74 &   21.30 &   60.68 &   0.00 &  1.00 &    100.00 &     0.00 &    26.90 &    64.66 &   77.52 &  0.29 & $ 29.4 \pm $ 34.4 \\ \cline{3-16}
	& BPCER &  8.70 & 7.63 & 11.03 &   11.76 &   10.27 &   8.85 &  8.63 &     10.53 &    11.60 &    10.99 &    10.31 &   10.23 &  7.10 & $  9.8 \pm $  1.4 \\ \cline{3-16}
	& ACER  & 13.94 & 8.30 &  6.38 &   16.53 &   35.47 &   4.43 &  4.81 &     55.27 &     5.80 &    18.95 &    37.48 &   43.87 &  3.69 & $ 19.6 \pm $ 17.4 \\ \cline{3-16}
	& EER   & 11.68 & 7.94 &  7.22 &   15.04 &   21.30 &   3.78 &  4.52 &     26.49 &     1.23 &    14.89 &    23.28 &   18.90 &  4.82 & $ 12.3 \pm $  8.2 \\ \midrule

    \multirow{4}{*}{MCCNN (BCE)}

    &APCER &   38.93 &  30.60 &   7.85 &   20.00 &  32.56 &   0.00 &   2.00 &  70.65 &   0.00 &  29.00 &   46.69 &  57.32 &  23.20 &  $27.6 \pm 22.1 $\\ \cline{3-16}
	&BPCER &    7.10 &   6.45 &   7.48 &   10.04 &  12.56 &   8.59 &  10.04 &   9.96 &  11.72 &  11.37 &   12.75 &   7.71 &   9.89 &  $ 9.6 \pm  2.0 $\\ \cline{3-16}
	&ACER  &   23.01 &  18.52 &   7.66 &   15.02 &  22.56 &   4.29 &   6.02 &  40.31 &   5.86 &  20.19 &   29.72 &  32.52 &  16.54 &  $18.6 \pm 11.1 $\\ \cline{3-16}
	&EER   &   17.08 &  11.83 &   7.56 &   12.82 &  16.09 &   0.71 &   6.85 &  25.94 &   2.29 &  16.30 &   18.90 &  22.82 &  13.13 &  $13.2 \pm  7.4 $\\ \midrule

	\multirow{4}{*}{\textbf{MCCNN (BCE+OCCL)-GMM}}
	&APCER &  11.79 &  9.53 &   3.12 &   3.70 &  39.20 &    0.00 &   3.12 &   44.57 &   0.00 &  21.60 &  19.34 &  35.55 &   0.00 & $ 14.7 \pm 15.9 $ \\ \cline{3-16}
	&BPCER &  13.44 & 16.15 &  16.26 &  20.23 &  11.11 &   13.74 &   8.66 &   15.23 &  12.67 &  10.42 &  14.31 &  18.40 &  27.33 & $ 15.2 \pm  4.8 $ \\ \cline{3-16}
	&ACER  &  12.61 & 12.84 &   9.69 &  11.97 &  25.16 &    6.87 &   5.89 &   29.90 &   6.34 &  16.01 &  16.83 &  26.97 &  13.66 & $ 14.9 \pm  7.8 $ \\ \cline{3-16}
	&EER   &  12.82 & 12.94 &  11.33 &  13.70 &  13.47 &    0.56 &   5.60 &   22.17 &   0.59 &  15.14 &  14.40 &  23.93 &   9.82 & $ \textbf{12.0} \pm  \textbf{6.9} $ \\

    \bottomrule

	\end{tabular}
	}
	}
\label{tab:siwm_results}
\end{table*}

Table \ref{tab:siwm_results} shows the performance of the proposed framework, again in RGB only scenario. CNN based methods performs much better than feature based methods in this case. It can be seen that the proposed approach achieves better performance as compared to baseline methods. The performance of the \textit{MCCNN (BCE+OCCL)-GMM} is better compared to the \textit{MCCNN(BCE)} model. It can be seen that the addition of the new loss function makes the classification of unseen attacks more accurate.

\subsection{Cross-database evaluations}

\begin{table}[h]

\begin{center}
\footnotesize
\caption{The results from the cross-database testing between WMCA and SiW-M datasets using the grandtest protocol, only RGB channels were used in this experiment.}
\label{tab:cross_test}

\resizebox{0.45\textwidth}{!}
{
{

\begin{tabular}{@{}c|c|c|c|c@{}}
\toprule
\multirow{2}{*}{Method} & \multicolumn{2}{c|}{\begin{tabular}[c]{@{}c@{}}trained on\\ WMCA\end{tabular}}                                     & \multicolumn{2}{c}{\begin{tabular}[c]{@{}c@{}}trained on\\ SiW-M\end{tabular}}                                       \\ \cline{2-5}
                        & \begin{tabular}[c]{@{}c@{}}tested on\\  WMCA\end{tabular} & \begin{tabular}[c]{@{}c@{}}tested on\\ SiW-M\end{tabular} & \begin{tabular}[c]{@{}c@{}}tested on\\ SiW-M\end{tabular} & \begin{tabular}[c]{@{}c@{}}tested on \\ WMCA\end{tabular} \\ \midrule
\rowcolor{Gray}
MC-RDWT-Haralick-SVM     &14.6      &29.6         & 15.1          &45.3                          \\
MC-LBP-SVM          &26.6      &45.5         & 19.6          &38.6        \\
\rowcolor{Gray}
MC-RDWT-Haralick-GMM     &27.9      &34.0         & 25.5          &43.6             \\
DeepPixBiS        &7.5       &49.1         & 14.7          &44.4       \\
\rowcolor{Gray}
MCCNN (BCE)       &12.1      &34.0         & 9.9           &42.3       \\ \midrule
\textbf{MCCNN (BCE+OCCL)-GMM}    &12.3      &31.9         &9.5            &41.8     \\ \bottomrule

\end{tabular}
}
}
\end{center}

\end{table}

As we could not perform cross-database evaluation between a multi-channel database and an RGB only database, we used only the RGB channels from two datasets for the cross-database evaluation. We have selected WMCA and SiW-M datasets as they are relatively large and consist of a wide variety of attacks.

From Table \ref{tab:cross_test}, it can be seen that the MCCNN model with and without the new loss performs comparably. In general, the performance in the cross-database setting is poor for all the models. The poor performance could be due to the disparity in acquisition conditions and the attack types. A wider variety of attacks makes it difficult for the classifier to identify attacks using RGB channels alone. The cross-database performance with this wide variety of attacks seems more challenging as compared to typical cross-database evaluations using only 2D attacks. Using multiple channels \cite{george_mccnn_tifs2019} may alleviate these issues. This also points to the limitation of RGB only methods while dealing with a wide variety of attacks.

\section{Discussions}

From the experiments in \textit{WMCA} database, it can be clearly seen that CNN based method outperforms the feature-based methods by a large margin. While comparing the \textit{MCCNN(BCE)} method to the proposed method, the performance is comparable in the known attack scenario. This indicates that the proposed One-Class GMM based approach performs par with binary classification, thanks to the embedding learned with the proposed loss function. Most of the approaches perform well in the \textit{unseen-2D} protocol since it can be clearly discriminated in many channels. Moreover, it shows that if the network is trained in challenging attacks, simpler attacks are easy to detect. While the performance is comparable in \textit{grandtest} and  \textit{unseen-2D} protocols, the proposed method achieves a large performance boost in the most challenging \textit{unseen-3D} protocol. The proposed loss function forces the network to learn a compact representation for \textit{bonafide} samples in the feature space. Both known and unknown attacks get mapped far from the \textit{bonafide} cluster in the feature space. The decision boundary learned by the one class model seems to be robust in identifying both seen and unseen attacks in such a scenario. This result is significant for several reasons. It is to be noted that in the \textit{unseen-3D} protocol, the network is trained with only 2D attacks, i.e., prints and replays. The proposed method achieves excellent performance in a test set consisting of challenging 3D attacks such as custom silicone masks, paper masks, mannequins, etc. The real-world implications of this approach is very promising. The proposed method can be used to develop robust PAD systems without the requirement of having to manufacture costly presentation attacks. The models can be trained on easy to obtain attacks based on availability. The proposed framework utilizes the available (known) attack categories to learn a robust representation to facilitate known and unseen attack detection. It is to be noted that the compact representation is made possible by the joint multi-channel representation used.

In practical deployment scenarios, it may not be possible to use all the four channels due to the computational or cost constraints. In such a situation, models trained on available channels can be selected based on the performance-cost ratio by sub selecting the channels. The results from the ablation study presented in Table \ref{tab:ablation_channels} can be used to determine which channels should be used in such cases.

Similarly, the experiments in \textit{MLFP} and \textit{SiW-M} database also shows CNN based methods outperform feature-based baselines. Although we have not used multi-channel information in the experiments, the experiment results showcase the performance improvement with the new loss function. Using the proposed framework together with network backbones designed specifically for RGB PAD might improve the results.

The cross-database performance shows the limitations of the RGB channel when tested with a wide variety of attacks. The performance of the baselines as well as the proposed approach is poor when only RGB data is used. This shows the challenging nature of RGB only PAD while considering a wide variety of attacks. The usage of multiple channels as done in \textit{WMCA} dataset might improve the performance.

\section{Conclusions}

Face presentation attack detection is often considered as a binary classification task which results in over-fitting to the known attacks leading to poor generalization against unseen attacks. In this paper, we address this problem with a new framework using a one-class classifier. A novel loss function is formulated, which forces the CNN to learn a compact yet discriminative representation for the face images. The \textit{bonafide} samples form a compact cluster in the feature space, thanks to the proposed loss function. A decision boundary around the \textit{bonafide} representation can be obtained using a one-class model. Both known and unknown attacks map far from the \textit{bonafide} cluster in the feature space which can be classified by the one-class model. The proposed framework introduces a new way to learn a robust PAD system from \textit{bonafide} and available (known) attack classes. The proposed system has been evaluated in the challenging \textit{WMCA}, \textit{MLFP}, and \textit{SiW-M} databases and was shown to outperform the baseline feature-based and CNN based methods in both known and unseen attack scenarios. The drastic improvement in the performance in \textit{unseen-3D} protocol in \textit{WMCA} shows the robustness of the proposed approach against unseen attacks, thanks to the multi-channel information. The proposed method also shows improvement even when used together with RGB channels alone. The source code and protocols to reproduce the results are made available publicly to enable further extensions of the proposed framework.

\section*{Acknowledgment}

Part of this research is based upon work supported by the Office of the
Director of National Intelligence (ODNI), Intelligence Advanced Research
Projects Activity (IARPA), via IARPA R\&D Contract No. 2017-17020200005.
The views and conclusions contained herein are those of the authors and
should not be interpreted as necessarily representing the official
policies or endorsements, either expressed or implied, of the ODNI,
IARPA, or the U.S. Government. The U.S. Government is authorized to
reproduce and distribute reprints for Governmental purposes
notwithstanding any copyright annotation thereon.

\ifCLASSOPTIONcaptionsoff
  \newpage
\fi

\bibliographystyle{IEEEtran}
\bibliography{refs_mcnn}

\begin{IEEEbiography}
  [{\includegraphics[width=1in,height=1.25in,clip,keepaspectratio]{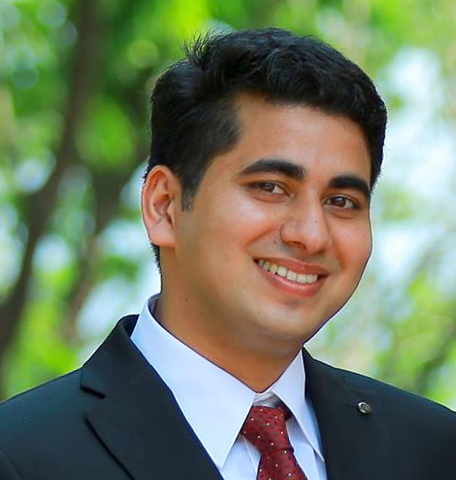}}]{Anjith George}
has received his Ph.D. and M-Tech degree from the Department of Electrical Engineering, Indian Institute of Technology (IIT) Kharagpur, India in 2012 and 2018 respectively. After Ph.D, he worked in Samsung Research Institute as a machine learning researcher. Currently, he is a post-doctoral researcher in the biometric security and privacy group at Idiap Research Institute, focusing on developing face presentation attack detection algorithms. His research interests are real-time signal and image processing, embedded systems, computer vision, machine learning with a special focus on Biometrics.
\end{IEEEbiography}
\begin{IEEEbiography}
 [{\includegraphics[width=1in,height=1.25in,clip,keepaspectratio]{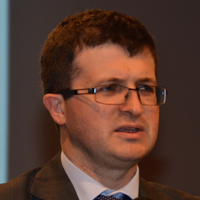}}]{S{\'e}bastien Marcel}
received the Ph.D. degree in signal processing from Universit\'{e} de Rennes I, Rennes, France, in 2000 at CNET, the Research Center of France Telecom (now Orange Labs). He is currently interested in pattern recognition and machine learning with a focus on biometrics security. He is a Senior Researcher at the Idiap Research Institute (CH), where he heads a research team and conducts research on face recognition, speaker recognition, vein recognition, and presentation attack detection (anti-spoofing). He is a Lecturer at the Ecole Polytechnique F\'{e}d\'{e}rale de Lausanne (EPFL) where he teaches a course on ``Fundamentals in Statistical Pattern Recognition.'' He is an Associate Editor of IEEE Signal Processing Letters. He has also served as Associate Editor of IEEE Transactions on Information Forensics and Security, co-editor of the ``Handbook of Biometric Anti-Spoofing,'' Guest Editor of the IEEE Transactions on Information Forensics and Security Special Issue on ``Biometric Spoofing and Countermeasures,'' and co-editor of the IEEE Signal Processing Magazine Special Issue on ``Biometric Security and Privacy.'' He was the Principal Investigator of international research projects including MOBIO (EU FP7 Mobile Biometry), TABULA RASA (EU FP7 Trusted Biometrics under Spoofing Attacks), and BEAT (EU FP7 Biometrics Evaluation and Testing).
\end{IEEEbiography}
\vfill
\end{document}